\documentclass[10pt,twocolumn,letterpaper]{article}

\usepackage[pagenumbers]{iccv} 

\definecolor{iccvblue}{rgb}{0.21,0.49,0.74}
\usepackage[pagebackref,breaklinks,colorlinks,allcolors=iccvblue]{hyperref}

\usepackage[accsupp]{axessibility}  

\usepackage{multirow}
\usepackage{hyperref}
\usepackage{amssymb}
\usepackage{pifont}
\usepackage{soul}
\usepackage{algorithm}
\usepackage{algpseudocode}
\usepackage{amsmath}
\usepackage{booktabs}
\usepackage{xcolor}
\usepackage{graphicx}
\usepackage{stfloats}
\definecolor{mygreen}{RGB}{0, 128, 0}
\usepackage{url}

\def\paperID{2587} 
\def\confName{ICCV}
\def\confYear{2025}

\title{UDC-VIT: A Real-World Video Dataset for Under-Display Cameras}

\author{
\centerline{\textbf{Kyusu Ahn}$^{3}$\thanks{\scriptsize{This work was done when he was a Ph.D. student at Seoul National University.}}~~~
\textbf{JiSoo Kim}$^{1}$~~~ 
\textbf{Sangik Lee}$^{4}$\thanks{\scriptsize{This work was done when he was an M.S. student at Seoul National University.}}~~~} \\
\centerline{\textbf{HyunGyu Lee}$^{2}$~~~
\textbf{Byeonghyun Ko}$^{2}$~~~ 
\textbf{Chanwoo Park}$^{2}$~~~ 
\textbf{Jaejin Lee}$^{1,2}$~~~}
\smallskip
\\
\small
$^1$Dept. of Data Science, Seoul National University, Seoul, Republic of Korea \\
\small
$^2$Dept. of Computer Science and Engineering, Seoul National University, Seoul, Republic of Korea \\
\small
$^3$Research Center, Samsung Display Co., Ltd., Yongin, Republic of Korea \\
\small
$^4$Mobile Display Electronics Development Team, Samsung Display Co., Ltd., Yongin, Republic of Korea \\
\small
$\{$kyusu.ahn, jisoo.kim, sangik85, devko, hyungyu, 99chanwoo, jaejin$\}$@snu.ac.kr \\ 
\centerline{\url{https://github.com/mcrl/UDC-VIT}}
}

\begin{document}
\maketitle
\begin{abstract}

Even though an Under-Display Camera (UDC) is an advanced imaging system, the display panel significantly degrades captured images or videos, introducing low transmittance, blur, noise, and flare issues. Tackling such issues is challenging because of the complex degradation of UDCs, including diverse flare patterns. However, no dataset contains videos of real-world UDC degradation. In this paper, we propose a real-world UDC video dataset called UDC-VIT. Unlike existing datasets, UDC-VIT exclusively includes human motions for facial recognition. We propose a video-capturing system to acquire clean and UDC-degraded videos of the same scene simultaneously. Then, we align a pair of captured videos frame by frame, using discrete Fourier transform (DFT). We compare UDC-VIT with six representative UDC still image datasets and two existing UDC video datasets. Using six deep-learning models, we compare UDC-VIT and an existing synthetic UDC video dataset. The results indicate the ineffectiveness of models trained on earlier synthetic UDC video datasets, as they do not reflect the actual characteristics of UDC-degraded videos. We also demonstrate the importance of effective UDC restoration by evaluating face recognition accuracy concerning PSNR, SSIM, and LPIPS scores. UDC-VIT is available at our official GitHub repository. 

\end{abstract}    
\section{Introduction}
\label{sec:intro}

\begin{figure*}[!t]
   \centering
   \includegraphics[width=0.65\linewidth]{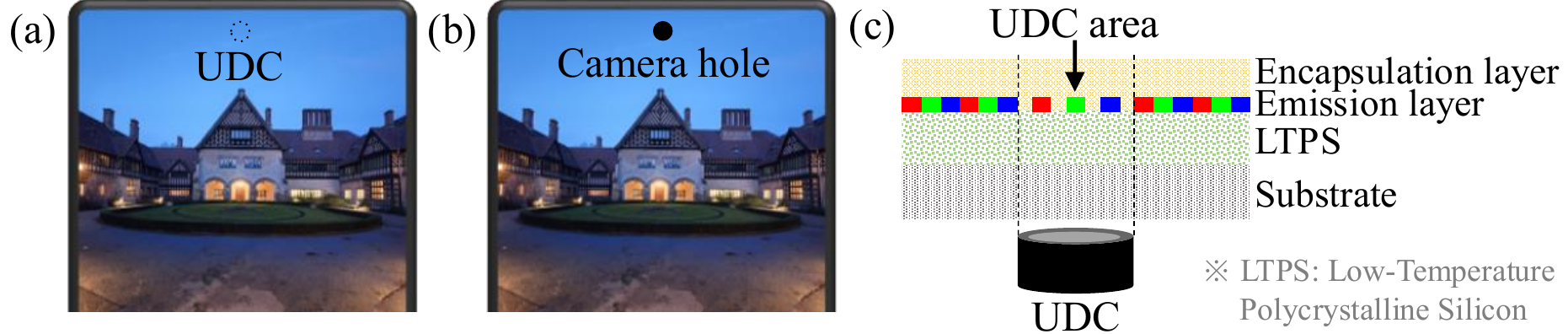}
   \caption{Comparison between under-display (UDC) and traditional hole-display cameras. (a) UDC. (b) Hole display camera. (c) The pixel structure of the UDC area. The UDC area exhibits a reduced pixel density due to the pixel pattern acting as diffraction slits.}
\label{fig:display_comparison}
\end{figure*}

Under-Display Camera (UDC) is an imaging system where the camera is positioned beneath the display. Modern smartphones, including the Samsung Galaxy Z-Fold series~\cite{samsungGalaxyZFold3,samsungGalaxyZFold4,samsungGalaxyZFold5,samsungGalaxyZFold6} and the ZTE Axon series~\cite{zteAxon20,zteAxon30,zteAxon40-ultra} have adopted UDCs. The UDC area, depicted in \cref{fig:display_comparison}, serves as display space under normal circumstances and acts as the light's passage to the camera when capturing pictures or videos. This design allows for a larger screen-to-body ratio, meeting the common consumer demand for a full-screen display without a camera hole or notch. However, UDC introduces severe and complex degradations such as reduced transmittance, noise, blur, and flare in a single image or video frame. Moreover, motion is also involved in UDC videos.

UDC degradation results from micrometer-scale diffraction of incoming light by display pixels~\cite{qin2016see}. To mitigate this, modern UDC smartphones adopt lower pixel density in the UDC area, as shown in \cref{fig:display_comparison}(c). However, this reduces the local display resolution and disrupts the natural video viewing experience, making high-quality video restoration essential.

Many studies have explored UDC image datasets, including synthetic datasets (\eg, T-OLED/P-OLED~\cite{zhou2021image} and SYNTH~\cite{feng2021removing}) and a pseudo-real dataset~\cite{feng2023generating}. Additionally, a real-world dataset called UDC-SIT~\cite{ahn2024udc} exists. 

Ahn~\etal~\cite{ahn2024udc} highlight the importance of training DNN models on real-world UDC datasets as synthetic datasets do not reflect the actual characteristics of UDC-degraded images. However, a real-world UDC video dataset and restoration model have yet to be introduced. Although several studies address the synthetic UDC video datasets~\cite{chen2023deep,liu2024decoupling}, they still fall short of accurately representing the actual properties of UDC videos. There are two main challenges in constructing a real-world UDC video dataset. One is to find a matching pair of UDC-distorted and ground-truth videos with high alignment accuracy. The other is to synchronize the time for all frames when capturing videos. 

This paper proposes a new UDC video dataset called UDC-VIT (\textbf{UDC}'s \textbf{VI}deo by \textbf{T}hunder Research Group). As far as we know, it is the first real-world UDC video dataset to address limitations of the existing UDC video datasets.

Using a non-polarizing cube beam splitter~\cite{thorlabs_beamsplitter}, we create a video-capturing system to minimize discrepancies between paired frames. We cut the UDC area of a smartphone display (\eg, Samsung Galaxy Z-Fold 5~\cite{samsungGalaxyZFold5}) and attach it to the beam splitter. Two Arducam Hawk-Eye (IMX686) camera modules~\cite{hawkeye} are placed on both sides of the beam splitter. These modules, operated by a Raspberry Pi 5~\cite{rpi5}, capture synchronized video frame pairs using the Message Passing Interface (MPI) barrier.

\cref{fig:capturing_system} shows our UDC video capturing system. Despite the meticulous design, inevitable pixel-position difference occurs. We correct this difference between the two matched frames for the same scene by using the DFT~\cite{brigham1988fast} following the previous work by Ahn~\etal~\cite{ahn2024udc}.

The contributions of this paper are as follows:
\begin{itemize}
\item We address the limitations of existing datasets, including unrealistic degradations, flares, and artifacts, emphasizing the need for a real-world dataset (\cref{sec:datasets_comparison} and Appendix~\ref{sec:vidudc33k_strange}).

\item We provide UDC-VIT, a real-world UDC video dataset that accurately reflects actual UDC degradations, ensuring precise spatial and temporal alignment through our meticulously designed video-capturing system (\cref{sec:dataset_acquisition} and Appendix~\ref{sec:add_info_udcvit}).

\item We describe UDC-VIT's effectiveness through extensive experiments, comparing it with an existing synthetic dataset using six deep-learning models (\cref{sec:experiments}, Appendix~\ref{sec:apndx_cross}, and Appendix~\ref{sec:analysis}).

\item We highlight the importance of restoring UDC degradation for applications like Face ID by evaluating face recognition accuracy at different restoration levels. Our dataset includes real-world face videos, enhancing its relevance for practical use (\cref{sec:experiments}).
\end{itemize}

\section{Related Work}
\label{sec:related_work}

\paragraph{Existing UDC image datasets.}
There has been extensive research on UDC still image datasets. Zhou~\etal~\cite{zhou2021image} propose the T-OLED/P-OLED datasets. Images are displayed on a monitor, and paired images are captured with and without a T-OLED/P-OLED display in front of the camera. However, due to the limited dynamic range of the monitor, flares are almost absent in their datasets. Feng~\etal~\cite{feng2021removing} propose the SYNTH dataset. They convolve the measured point spread function (PSF) of ZTE Axon 20~\cite{zteAxon20} with clean images~\cite{hdrihaven}, exhibiting flares. However, it has limitations such as the absence of noise and \textit{spatially variant flares}. Notably, UDC distortion gradually increases from the center of the camera lens to outwards, leading to spatially distorted flares~\cite{yoo20227}. Feng~\etal~\cite{feng2023generating} propose a pseudo-real dataset by capturing paired images of similar scenes using two cameras (\eg, ZTE Axon 20 UDC~\cite{zteAxon20} and iPhone 13 Pro camera~\cite{iPhone13Pro}). However, they use two cameras, leading to geometric misalignment. They improve the geometric misalignment using AlignFormer~\cite{feng2023generating}. Nonetheless, they encounter challenges with alignment accuracy. Ahn~\etal propose a real-world dataset called UDC-SIT and an image-capturing system~\cite{ahn2024udc,ahn2025device}. They attach Samsung Galaxy Z-Fold 3~\cite{samsungGalaxyZFold3}'s UDC area to a lid. Paired images are acquired by opening and closing the lid onto the Samsung Galaxy Note 10's standard camera~\cite{samsungGalaxyNote10}. They use DFT to align the misalignment between the paired images that occurs during the opening and closing of the lid~\cite{ahn2025image}. The images in the UDC-SIT dataset contain the actual UDC degradation (\eg, \textit{spatially variant flares}). Finally, Wang~\etal~\cite{wang2024lrdif} and Tan~\etal~\cite{tan2023blind} propose face datasets for facial expression recognition or restoration. They are synthesized using a GAN-based model trained on the T-OLED/P-OLED datasets~\cite{zhou2021image}, which lacks realistic UDC degradation, particularly flares. Moreover, the datasets are not publicly available.

\paragraph{Existing UDC video datasets.}
Research has been conducted on synthetic UDC video datasets. Chen~\etal~\cite{chen2023deep} propose the PexelsUDC-T/P datasets. They train a GAN-based UDC video generation model using T-OLED/P-OLED datasets~\cite{zhou2021image}, which do not show flares. They generate UDC-degraded videos using clean videos~\cite{pexels}. The datasets are not publicly available. Liu~\etal~\cite{liu2024decoupling} propose the VidUDC33K dataset. They convolve the measured PSF on the clean video frames~\cite{hdrihaven} to show flares. They simulate the dynamic change of the PSF~\cite{kwon2021controllable} between consecutive frames following the previous work~\cite{babbar2022homography, liu2022unsupervised, ye2021motion}. However, flares in their dataset are unrealistic.

\paragraph{UDC image restoration.}
There has been active research on UDC image restoration. DISCNet~\cite{feng2021removing} incorporates the domain knowledge of the image formation model. UDC-UNet~\cite{liu2022udc}, a second performer of MIPI challenge~\cite{feng2022mipi}, introduces kernel branches to incorporate prior knowledge and condition branches for spatially variant manipulation. SFIM~\cite{ahn2025integrating} integrates spatial and frequency information to effectively address flare artifacts in real UDC images.

\paragraph{Video restoration.}
Many studies have focused on video restoration models for general tasks, such as denoising~\cite{tassano2020fastdvdnet}, deblurring~\cite{wang2019edvr,zhong2020efficient}, and super-resolution~\cite{wang2019edvr}. Unlike image restoration, which only focuses on a spatial dimension, video restoration leverages temporal information. FastDVDNet~\cite{tassano2020fastdvdnet} uses a two-step denoising process in a multi-scale architecture to leverage temporal information without explicit motion estimation. EDVR~\cite{wang2019edvr} aligns features using deformable convolutions~\cite{dai2017deformable} and applies both temporal and spatial attention to highlight essential features. ESTRNN~\cite{zhong2020efficient} integrates residual dense blocks into RNN cells for spatial feature extraction and employs a spatiotemporal attention module for feature fusion. However, studies on UDC video restoration are still rare. DDRNet~\cite{liu2024decoupling}, the pioneering work to address UDC video degradation, adopts a recurrent architecture that merges multi-scale feature learning and bi-directional propagation.
\section{Dataset Acquisition}
\label{sec:dataset_acquisition}
Since obtaining well-synchronized and precisely aligned paired videos for the same scene is challenging, we carefully design both hardware and software for video capture.

\subsection{The Video Capturing System}
\label{sec:capturing_system}
\begin{figure}[b]{
    \centering
    \begin{minipage}[t]{0.99\linewidth}
        \begin{minipage}[t]{0.48\linewidth}
            \centering
            \includegraphics[width=\linewidth, height=100pt]{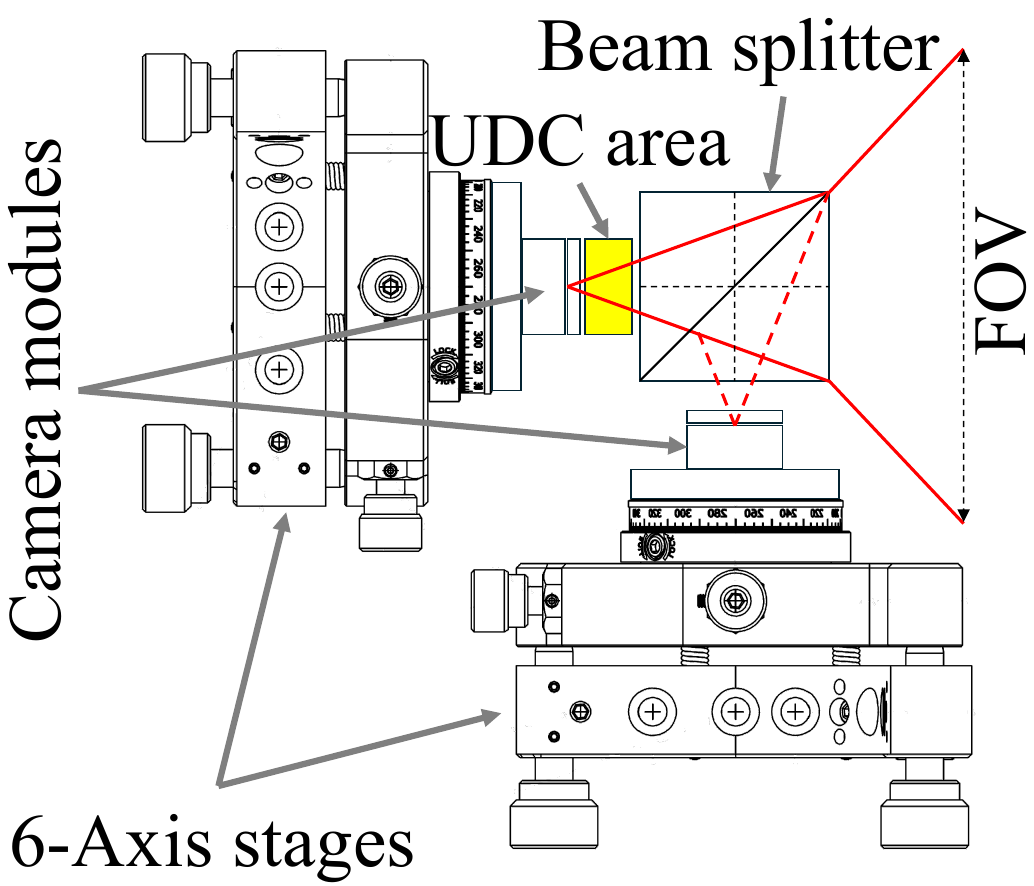}
            (a)
        \end{minipage}
        \hfill
        \begin{minipage}[t]{0.44\linewidth}
            \centering
            \includegraphics[width=\linewidth, height=100pt]{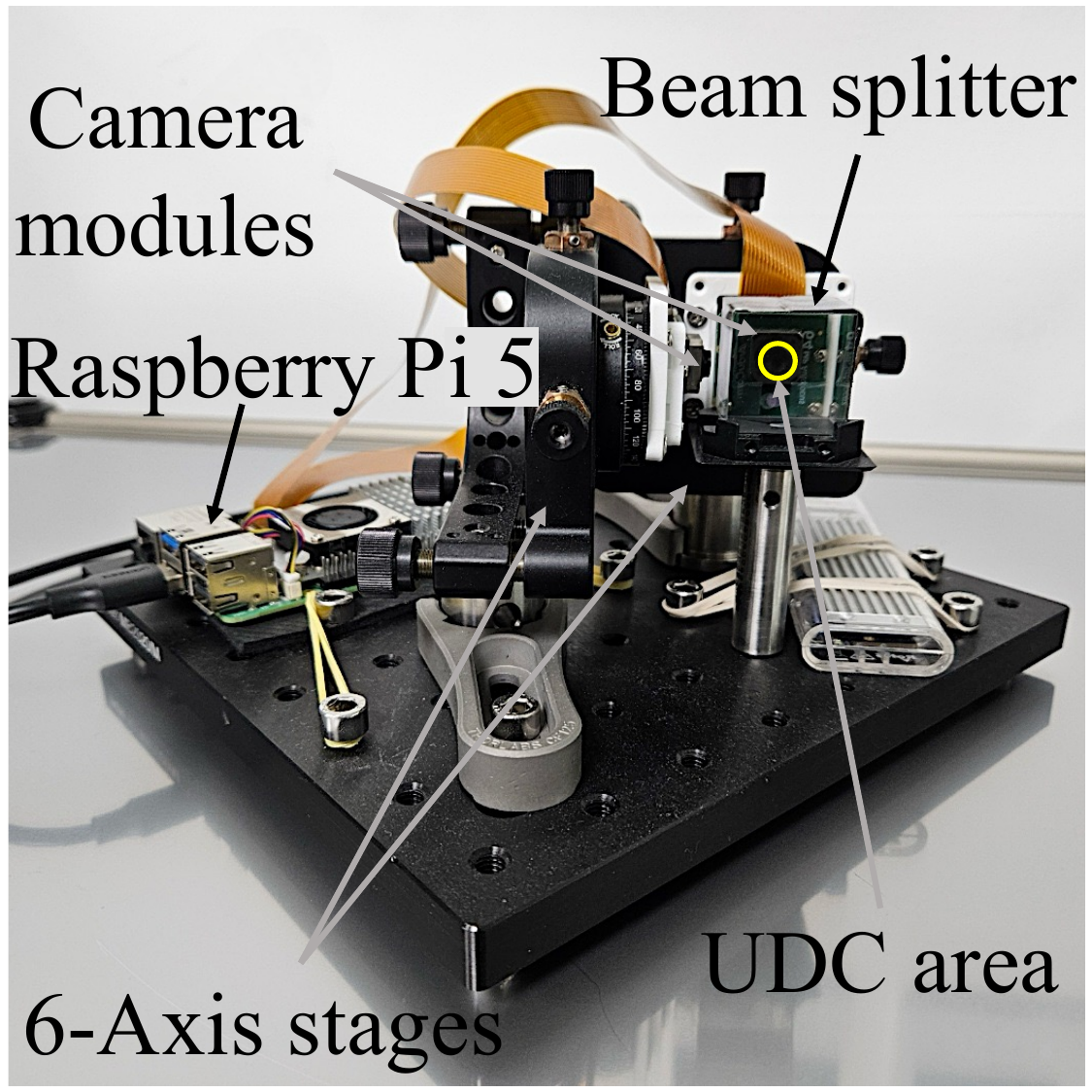}
            (b)
        \end{minipage}
    \end{minipage}
    \caption{The UDC video-capturing system. (a) The optical layout of the dual camera combiner. The UDC area is enlarged for a better view. (b) The proposed video-capturing system.}
    \label{fig:capturing_system}}
\end{figure}

As shown in \cref{fig:capturing_system}, we present a UDC video capturing system with two camera modules, a display panel for the UDC area, a beam splitter, two 6-axis stages, and a single-board computer. In this setup, one of the two camera modules is in low light conditions caused by the display panel, making synchronization between paired frames more challenging than in previous beam splitter setups~\cite{hwang2015multispectral,joze2020imagepairs,li2023real,rim2020real}. To capture synchronized videos for the same scene, we propose a UDC video-capturing system that ensures precise camera synchronization and frame alignment.

\begin{figure}[t]
   \centering
   \includegraphics[width=0.95\linewidth]{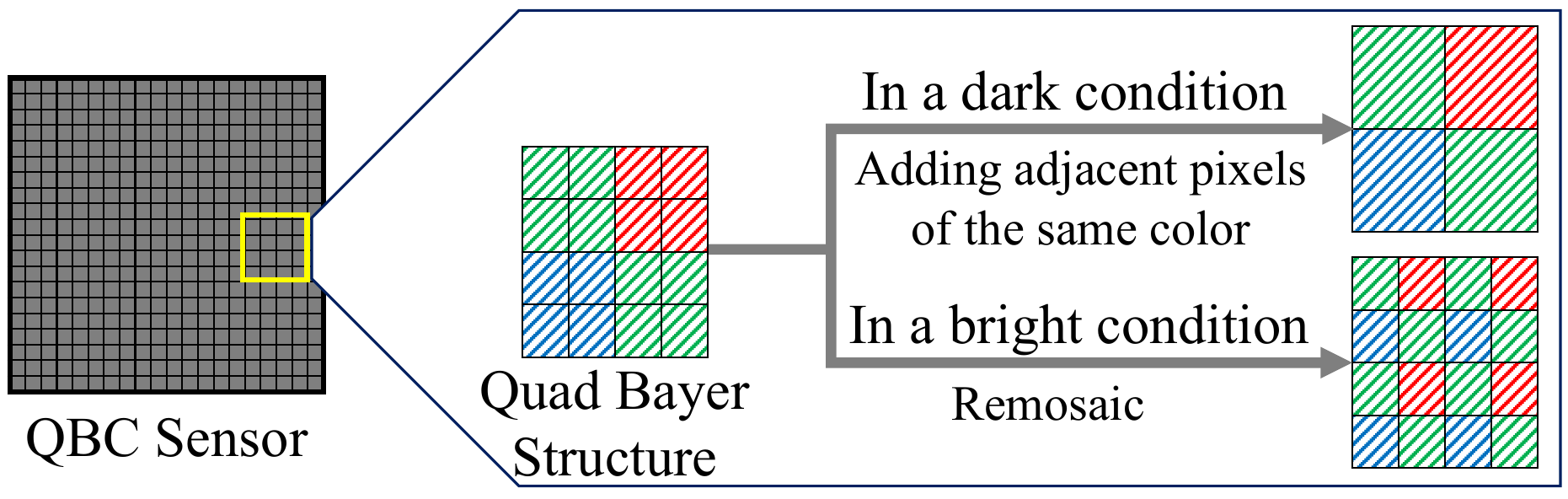}
   \caption{Quad Bayer Coding (QBC).}
\label{fig:qbc_concept}
\end{figure}

\paragraph{The camera module.}
We use the Hawk-Eye (IMX686)~\cite{hawkeye} to ensure that UDC-VIT exhibits a similar UDC degradation as Samsung Galaxy Z-Fold 5's UDC. Both devices use \textit{Quad Bayer Coding (QBC)}, a technique designed to mitigate the reduced sensor luminance sensitivity often associated with higher camera resolutions or smaller sensor pixel sizes~\cite{qbc_sony,qbc_wiki}. As shown in \cref{fig:qbc_concept}, four adjacent pixels share the same color filter in the quad Bayer structure. These pixels are grouped to increase sensitivity and reduce noise in low-light conditions (\eg, in the UDC setting). Conversely, in bright conditions, the sensor reverts the pixels to the Bayer structure through the remosaicing process, maintaining the Bayer sensor's high resolution. We apply pixel binning and ISP to both GT and degraded captures to ensure resolution consistency and synchronization via stable auto-exposure in well-lit conditions. Please see Appendix~\ref{sec:add_info_udcvit} for details.

\paragraph{The beam splitter.}
We use a non-polarizing cube-shaped beam splitter (\eg, Thorlabs CCM1-BS013~\cite{thorlabs_beamsplitter}) to enable the two cameras to capture the same scene. The beam splitter divides the incident light into two directions with a beam deviation of $0 \pm 5$ arcminutes at a 50:50 ratio. The unused optical path is black-coated to minimize image contrast loss due to scattering. They can capture the same scene by aligning the two cameras to the beam splitter's split fields of view (FOV). \cref{fig:capturing_system}(a) illustrates the optomechanical configuration of our beam splitter–based dual-camera setup.

\paragraph{Kinematic optical mount.}
Despite many studies using beam splitters for paired image dataset collection~\cite{hwang2015multispectral,joze2020imagepairs,li2023real,rim2020real}, this paper is the first to apply them in UDC research. It presents challenges to align the optical paths of the display panel's UDC area, a beam splitter, and two camera modules. To ensure alignment between the cameras' optical axes and the beam splitter, we employ Thorlabs K6XS 6-axis kinematic optical mounts~\cite{thorlabs_6_axis}. Each camera module is mounted on a K6XS mount, allowing for shifts, rotations, and tilts across the six axes to align their FOV.

\paragraph{The controller.}
We use a Raspberry Pi 5~\cite{rpi5} that has two four-lane MIPI interface connections for high bandwidth to synchronize the two high-resolution cameras. It ensures stable high-resolution video recording.
To synchronize the cameras, we use independent streamers managed by MPI barriers~\cite{mpi2023}, achieving synchronization with an accuracy margin of up to 8 msec. Consistent frame rates for both cameras are ensured using the uncompressed binary dumps (\eg, \texttt{YUV420} format). Despite these settings resulting in less than a 0.5 fps (8 msec) difference between paired frames, rapid movements may still cause the cameras to capture different scenes. Thus, videos capturing such objects (\eg, speeding cars) are excluded from the dataset.

\subsection{Obtaining Aligned Video Pairs}
\label{sec:align_vit}

This section explains optical axis and FOV alignment, criteria for FOV alignment, and test cases. As illustrated in \cref{alg:alignment_process}, using a real-time viewing system, we roughly align the cameras, fine-tune them with K6XS and DFT, evaluate accuracy (\eg, PCK), record the video, and manually select the final footage. For PCK details, see \cref{para:pck}.

\begin{algorithm}[h]
\caption{Aligned video capturing process.}
\label{alg:alignment_process}
\centering
\begin{minipage}{1.0\linewidth}
\begin{small}
\begin{algorithmic}
\Ensure Paired frame alignment accuracy exceeds 90\%.
\While{Average PCK ($\alpha = 0.005$) $<$ 90\%}
  \State \textbf{Initial setup.} Adjust camera positions and beam splitter for roughly similar views.
  \State \textbf{Fine-tuning.} Precisely adjust rotation/tilt angles and horizon/vertical position of K6XS by observing a $12 \times 9$ checkerboard and everyday scenes in the live view system.
  \State \textbf{Alignment and evaluation.} Align paired frames using DFT and compute the average PCK.
\EndWhile
  \State \textbf{Video recording.} Capture paired videos for the same scene.
  \State \textbf{Final selection.} Retain only aligned and synchronized videos approved by all authors.
\end{algorithmic}
\end{small}
\end{minipage}
\end{algorithm}

\paragraph{DFT alignment.}
Despite the careful design of our video-capturing system, misalignments, such as \textit{shifts}, \textit{rotations}, and \textit{tilts}, still occur between paired frames. Previous alignment methods, such as SIFT~\cite{lowe2004distinctive}, RANSAC~\cite{fischler1981random}, and deep learning approaches~\cite{feng2023generating}, struggle to perform well in the existence of severe degradation introduced by the UDC. Thus, we use DFT to achieve degradation-resilient alignment~\cite{ahn2024udc}. Please see Appendix~\ref{sec:add_info_udcvit} for details.

The alignment process is summarized as \textit{shift, rotate, and crop} paired frames using DFT. Captured videos have an original frame size of $(1920, 1080, 3)$. The ground-truth frame is center-cropped to $(1900, 1060, 3)$, and the degraded frame undergoes a cropping around the center. To align the cropped degraded frame $\mathcal{D}$ with the cropped ground-truth frame $\mathcal{G}$, we iteratively \textit{shift} the $(x, y)$ coordinates and \textit{rotate} the frames to find the point of minimum loss. Our focus is on addressing \textit{shifts} and \textit{rotations} while excluding \textit{tilts}. Handling \textit{tilts} is challenging because of the need for perspective transforms optimized for objects in the same plane within a single image. Despite not considering \textit{tilts}, our video-capturing system minimizes all \textit{shifts}, \textit{rotations}, and \textit{tilts} so that they do not significantly affect the alignment, as confirmed by our experiment (the PCK values in \cref{tab:align_pck}). The loss function $\mathcal{L}$ for the alignment between $\mathcal{D}$ and $\mathcal{G}$ is defined as below:
\begin{small}
\begin{equation}
 \begin{split}
 \mathcal{L} = & \lambda_1 \sum\limits_{x=0}^{M-1}\sum\limits_{y=0}^{N-1} (\mathcal{D}(x,y) - \mathcal{G}(x,y))^2  \\
 & + \lambda_2 \sum\limits_{u=0}^{M-1}\sum\limits_{v=0}^{N-1} \Delta \mathcal{F}_{amp}(u,v) + \lambda_3 \sum\limits_{u=0}^{M-1}\sum\limits_{v=0}^{N-1} \Delta \phi (u,v),
 \end{split}
 \label{eq:loss}
\end{equation}
\end{small}
where the first term is the mean squared error, $\Delta \mathcal{F}_{amp}(u,v)$ and $\Delta \phi(u,v)$ represent the L1 distance for the amplitude and phase, respectively. They are defined as $\Delta \mathcal{F}_{amp}(u,v) = |\mathcal{F}_\mathcal{D}(u,v) - \mathcal{F}_\mathcal{G}(u,v)|$ and $\Delta \phi(u,v) = |\phi_\mathcal{D}(u,v) - \phi_\mathcal{G}(u,v)|$. Note that $\mathcal{F}(u,v)$ is the frequency value at the point $(u,v)$ in the frequency domain. Following Ahn et al.'s setting~\cite{ahn2024udc}, we use $\lambda_1 = \lambda_3 = 1, \lambda_2 = 0$. See Appendix~\ref{sec:add_info_udcvit} for the alignment algorithm.

\section{Comparison with the Existing UDC Datasets}
\label{sec:datasets_comparison}

Many synthetic UDC datasets, including VidUDC33K~\cite{liu2024decoupling}, formulate the UDC degradation as follows:
\begin{equation}
  \label{eq:udc_formulation}
  I_t^{D} = f(\gamma \cdot I_{t}^{G} * k_t + n),
\end{equation}
where $I_t^{D}$ and $I_{t}^{G}$ are the UDC-degraded and ground-truth frames, respectively. $\gamma$ is the intensity scaling factor, $k_t$ refers to the diffraction kernel (i.e., PSF), $n$ is the noise, and $f$ denotes the clamp function for the pixel value saturation. 

Ideally, we would like to compare UDC-VIT with two existing UDC video datasets, PexelsUDC-T/P~\cite{chen2023deep} and VidUDC33K~\cite{liu2024decoupling}. However, since PexelsUDC-T/P is not publicly available, we use the P-OLED dataset~\cite{zhou2021image}, which is used to train the GAN model that generates synthetic PexelsUDC-T/P videos. \cref{tab:datasets_comparison} gives a summary of the nine UDC datasets. The resolution and frame rate (fps) of UDC-VIT are FHD and 60 fps, respectively, in line with the Samsung Galaxy Z-Fold 5's specifications.

\begin{table*}[!t]
 \caption{Comparison of the UDC datasets. The dataset size refers to the number of images in the image dataset or the total number of frames in the video dataset, calculated as the product of the number of video clips and the number of frames per clip. For example, the UDC-VIT dataset consists of 647 video clips with 180 frames per clip, so the total number of frames is 116,460.}
  \label{tab:datasets_comparison}
  \begin{center}
  \resizebox{0.9\linewidth}{!}{%
  \begin{tabular}{l|ccccccccc}
    \toprule
    \multirow{2}{*}{Dataset} & \multirow{2}{*}{Type} & \multirow{2}{*}{Scene} & \multirow{2}{*}{Dataset size} & \multirow{2}{*}{Resolution} & \multirow{2}{*}{fps} & Flare    & Face & Publicly & \multirow{2}{*}{Publication} \\
                             &  &  &  &  &  & presence & recognition & available & \\
    \toprule
    T/P-OLED~\cite{zhou2021image}    & Image & Synthetic     & 300                       & $1024 \times 2048 \times 3$ & -       &           &           & \ding{52} & CVPR '21\\
    \midrule
    SYNTH~\cite{feng2021removing}         & Image & Synthetic     & 2,376                     & $800  \times 800  \times 3$ & -       & \ding{52} &           & \ding{52} & CVPR '21 \\
    \midrule
    Pseudo-real~\cite{feng2023generating} & Image & \textbf{Real} & 6,747                     & $512  \times 512  \times 3$ & -       & \ding{52} &           & \ding{52} & CVPR '23 \\
    \midrule
    UDC-SIT~\cite{ahn2024udc}             & Image & \textbf{Real} & 2,340                     & $1792 \times 1280 \times 4$ & -       & \ding{52} &           & \ding{52} & NeurIPS '23 \\
    \midrule
    Tan~\etal~\cite{tan2023blind}   & Image & Synthetic     & 73,000                    & -                           & -       &           & \ding{52} &           & TCSVT '23 \\
    \midrule
    Wang~\etal~\cite{wang2024lrdif}   & Image & Synthetic     & 56,126                    & -                           & -       &           & \ding{52} &           & ICIP '24 \\
    \midrule
    \multirow{2}{*}{PexelsUDC-T/P~\cite{chen2023deep}} & \multirow{2}{*}{\textbf{Video}} & \multirow{2}{*}{Synthetic} & $160 \times 100$          & \multirow{2}{*}{$1280 \times 720  \times 3$} & \multirow{2}{*}{25-50}   &           &           &        & \multirow{2}{*}{arXiv '23}   \\
                                                       &       &           & ($16,000$) \\
    \midrule
    \multirow{2}{*}{VidUDC33K~\cite{liu2024decoupling}}    & \multirow{2}{*}{\textbf{Video}} & \multirow{2}{*}{Synthetic}     & $677 \times 50$   & \multirow{2}{*}{$1920 \times 1080 \times 3$} &    \multirow{2}{*}{-}     & \multirow{2}{*}{\ding{52}} &           & \multirow{2}{*}{\ding{52}} & \multirow{2}{*}{AAAI '24} \\
                                                       &       &           & ($33,850$) \\
    \midrule
    \multirow{2}{*}{UDC-VIT}                               & \multirow{2}{*}{\textbf{Video}} & \multirow{2}{*}{\textbf{Real}} & \textbf{$647 \times 180$} & \multirow{2}{*}{$1900 \times 1060 \times 3$} & \multirow{2}{*}{60}      & \multirow{2}{*}{\ding{52}} & \multirow{2}{*}{\ding{52}} & \multirow{2}{*}{\ding{52}} & \multirow{2}{*}{ICCV '25} \\
                                                       &       &           & ($116,460$) \\
    \bottomrule
  \end{tabular}
  }
  \end{center}
\end{table*}

\begin{figure}[!b]
\begin{minipage}{\linewidth}
   \centering
   \includegraphics[width=\linewidth]{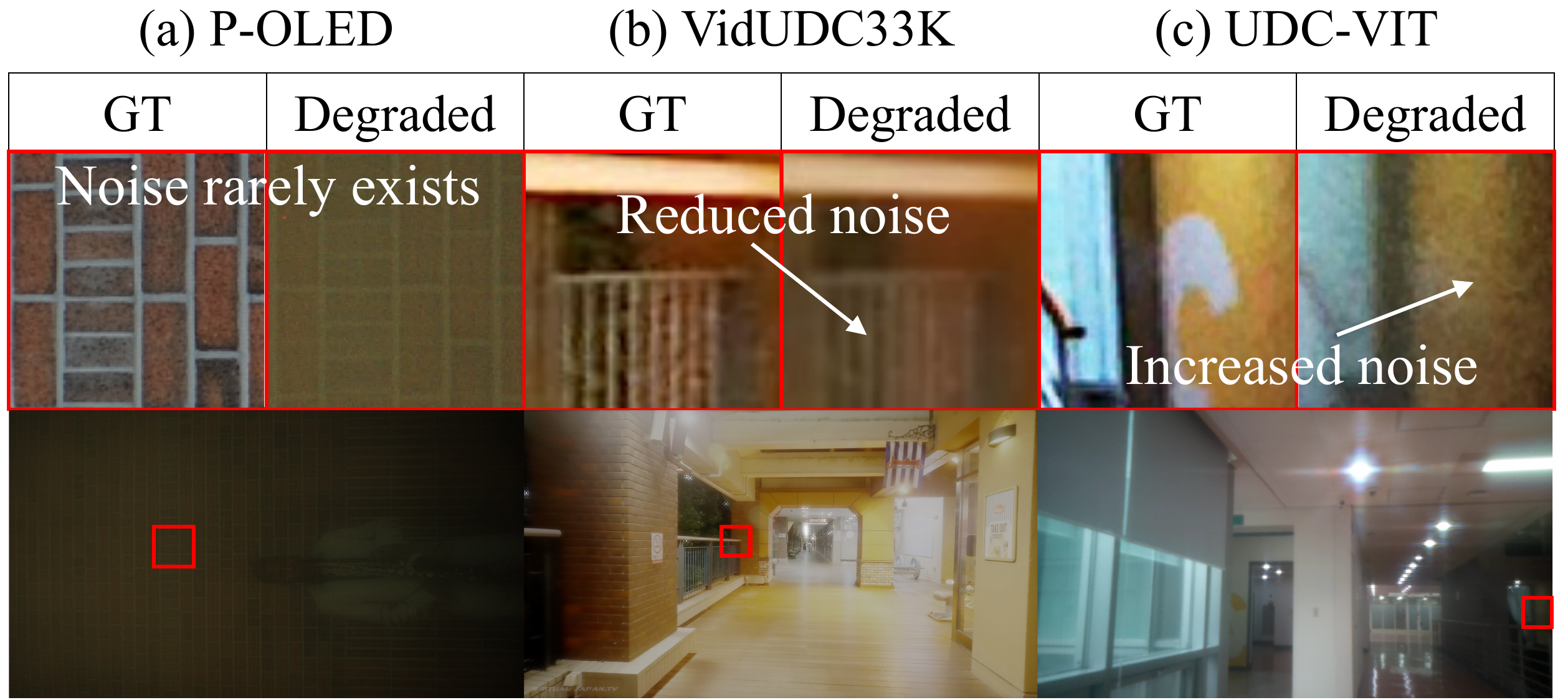}
   \caption{Comparison of the decrease in transmittance and digital noise by the UDC. (a) P-OLED dataset rarely depicts noise. (b) In the VidUDC33K dataset, the degraded frame decreases digital noise compared to the ground truth (GT) frame. (c) UDC-VIT dataset illustrates an increase in digital noise in the degraded frame. The brightness has been adjusted to improve visibility.}
\label{fig:trans_noise}
\end{minipage}
\end{figure}

\paragraph{Noise and transmittance decrease.}
The camera sensor amplifies the desired signal and unwanted noise in low-light conditions. In the UDC setting, where the sensor is beneath the display panel, the transmittance decreases, leading to amplified noise. The camera sensors with QBC, used in the Samsung Galaxy Z-Fold series (related to UDC-VIT)~\cite{samsungGalaxyZFold3,samsungGalaxyZFold4,samsungGalaxyZFold5,samsungGalaxyZFold6} and ZTE Axon series (related to VidUDC33K)~\cite{zteAxon20,zteAxon30,zteAxon40-ultra}, can influence the noise pattern and pixel intensity~\cite{qbc_sony}. Thus, simply adding noise and adjusting the intensity scaling values in \cref{eq:udc_formulation} may not accurately depict real-world noise and transmittance reduction. For example, in the VidUDC33K dataset, the degraded frame's noise level is somewhat lower than the ground truth, as shown in \cref{fig:trans_noise}(b). Similarly, the P-OLED dataset, captured in a controlled setting, exhibits unrealistic noise and excessive transmittance decrease, as depicted in \cref{fig:trans_noise}(a). In contrast, UDC-VIT in \cref{fig:trans_noise}(c) accurately shows actual transmittance decrease and digital noise resulting from quantizing digital image signals. 


\begin{figure}[!b]
   \centering
   \includegraphics[width=1.00\linewidth]{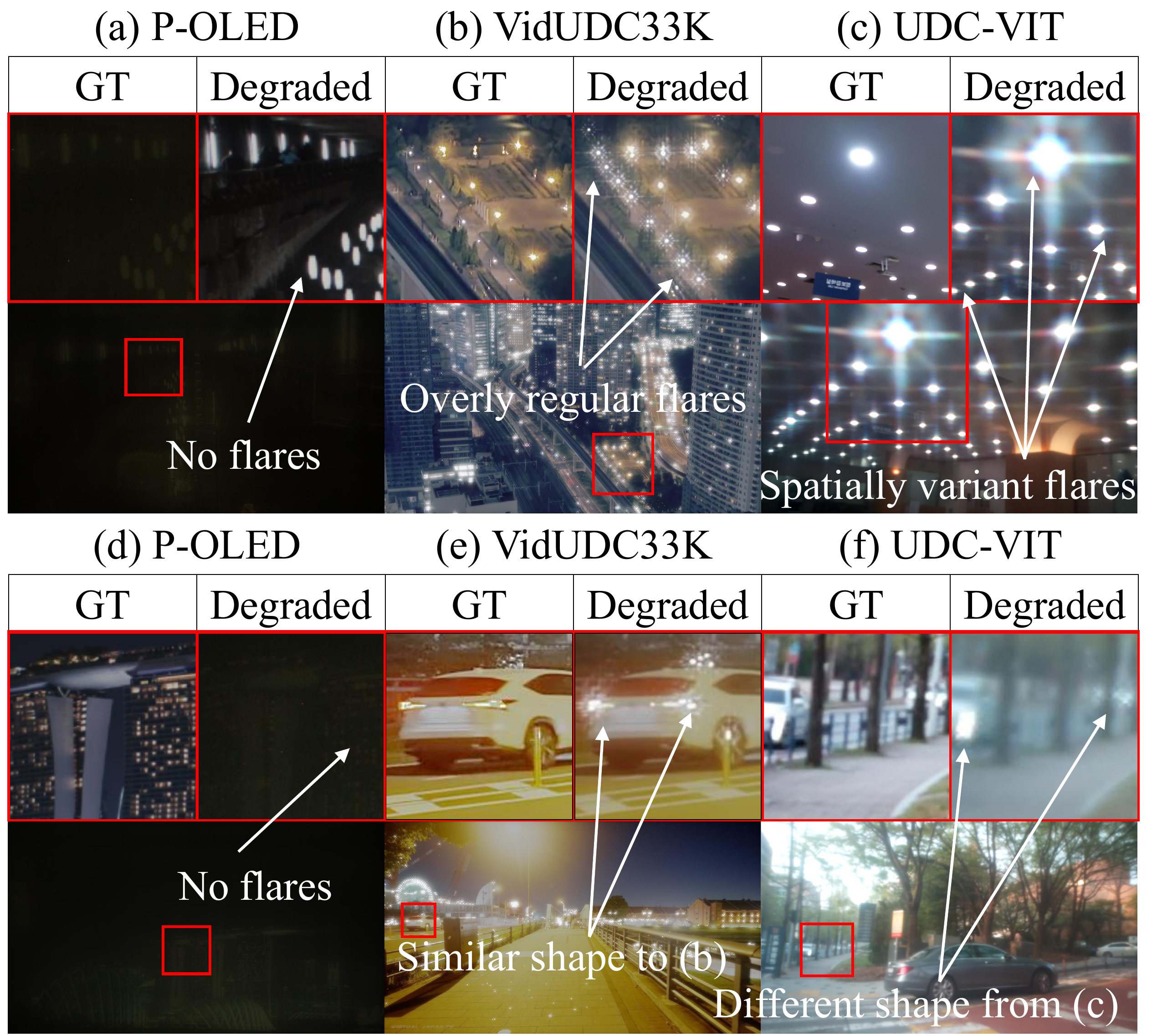}
   \caption{Comparison of flares. P-OLED shows no flares ((a) and (d)). VidUDC33K displays overly regular flares and light source invariant flares ((b) and (e)). In contrast, UDC-VIT uniquely presents \textit{spatially variant flares} and \textit{light source variant flares} ((c) and (f)).}
\label{fig:light_spatial_flare}
\end{figure}

\paragraph{Flares.}
Conventional lens flares stem from intense light scattering or reflection within an optical system~\cite{dai2022flare7k,dai2023nighttime}. In contrast, UDC flares arise from light diffraction as it passes through the display panel above the digital camera lens. Thus, it is crucial for each frame in the UDC video dataset to precisely depict the unique flare characteristics, including \textit{spatially variant flares}, \textit{light source variant flares}, and \textit{temporally variant flares}. The P-OLED dataset rarely exhibits flares as it captures images displayed on a monitor in a controlled environment (\cref{fig:light_spatial_flare}(a) and (d)).

Since UDC distortion increases outward from the camera lens center, \textit{spatially variant flares} manifest within an image~\cite{yoo20227}. Distorted PSFs must be convolved across different image regions to depict this flare distortion accurately. However, VidUDC33K applies the same PSF convolution across all areas using \cref{eq:udc_formulation}, failing to represent spatially variant flares, as illustrated in \cref{fig:light_spatial_flare}(b) and (e). Conversely, UDC-VIT effectively captures spatially variant flares (\cref{fig:light_spatial_flare}(c)).

\begin{figure}[!t]
   \centering
   \includegraphics[width=1.00\linewidth]{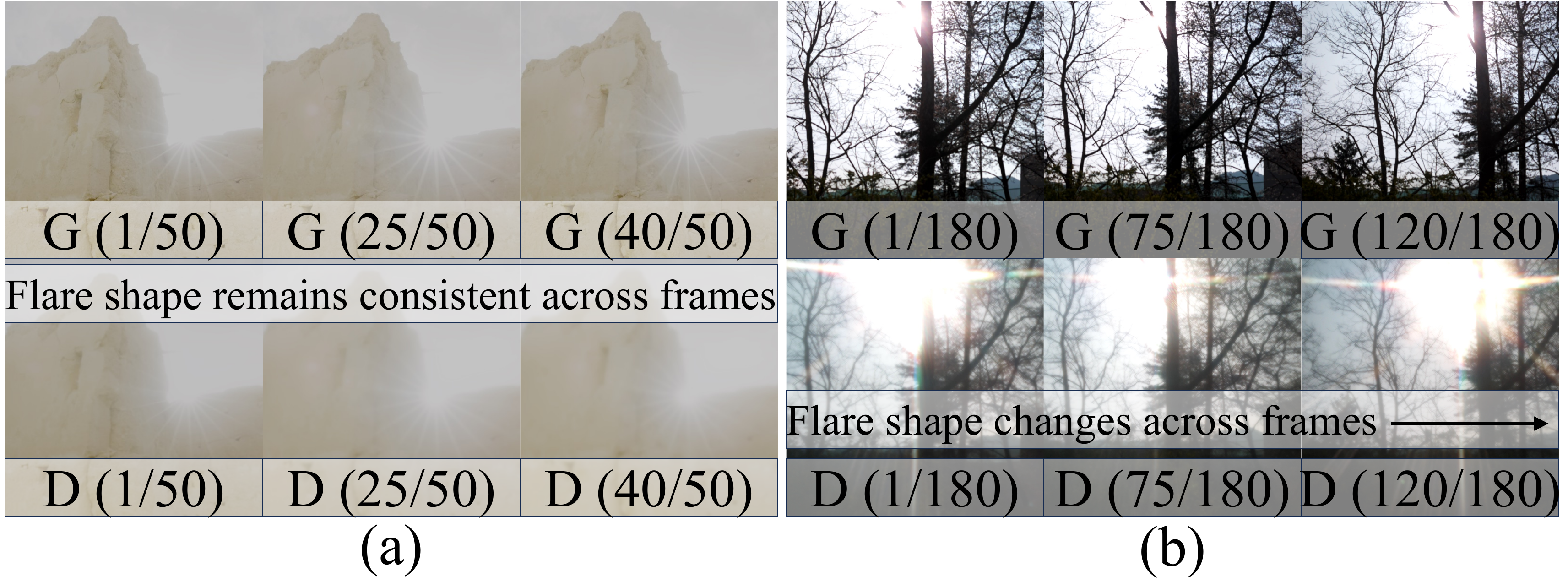}
   \caption{Temporally variant flares. Unlike (a) VidUDC33K, (b) UDC-VIT shows temporally variant flares. G and D are the ground truth and degraded frames, respectively. The numbers in parentheses represent \texttt{(the current frame number / the total number of frames)}.} 
\label{fig:temporal_flare}
\end{figure}

Various light sources, such as artificial (\eg, LED and halogen) and natural light, can alter the spectra, affecting UDC flares' shapes. However, VidUDC33K fails to depict \textit{light source variant flares}. As seen in \cref{fig:light_spatial_flare}(b) and (e), flare shapes remain similar despite different light sources. Conversely, UDC-VIT exhibits diverse flare shapes, as shown in \cref{fig:light_spatial_flare}(c) and (f) and \cref{fig:temporal_flare}(b).

A notable characteristic of UDC videos is \textit{temporally variant flares} caused by the camera's motion when capturing light sources. The motion results in changes in PSFs~\cite{kwon2021controllable}. However, in the VidUDC33K dataset, attempts to simulate PSF changes through inter-frame homography matrix computations using the method proposed by the previous studies~\cite{babbar2022homography, liu2022unsupervised, ye2021motion} yield rare \textit{temporally variant flares}, as shown in \cref{fig:temporal_flare}(a). Moreover, the shape of typical lens flares in ground-truth frames remains unchanged in degraded frames, indicating the failure of PSF convolution to replicate natural sunlight flares. Conversely, UDC-VIT effectively captures temporally varying flares (\cref{fig:temporal_flare}(b)). 

\begin{figure}[!t]
   \centering
   \includegraphics[width=1.00\linewidth]{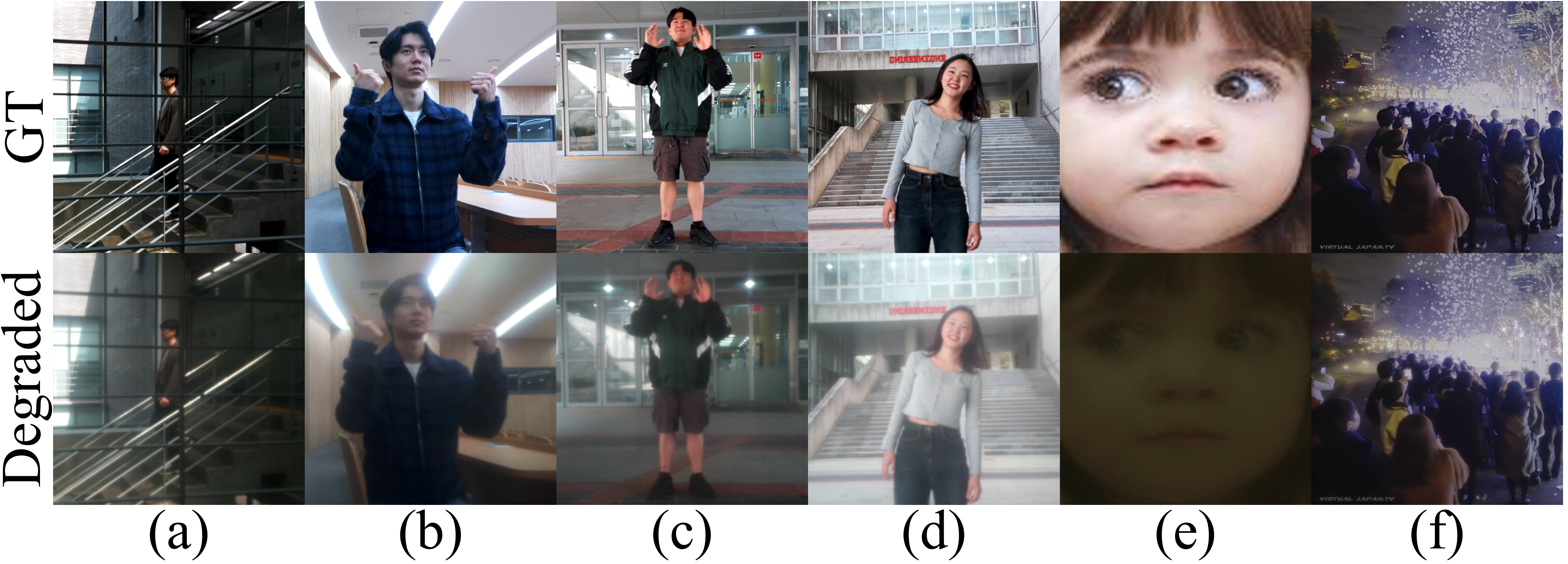}
   \caption{UDC-VIT features human motions, including (a) walking, (b) thumbs-up, (c) hand waving, and (d) body swaying. In contrast, Wang~\etal~\cite{wang2024lrdif}'s synthetic still image datasets for FER do not show the actual UDC degradations, as shown in (e). Moreover, it is not publicly available. VidUDC33K dataset includes humans but is limited to rear views, as shown in (f).}
\label{fig:face_recognition}
\end{figure}

\paragraph{Face recognition.}
\label{sec:face_recognition}
UDC-VIT stands out from other datasets by featuring videos tailored for face recognition (FR). Specifically, we focus on a \textit{pair-matching} task to identify if two face images belong to the same person. Some datasets, such as T-OLED/P-OLED, SYNTH, and VidUDC33K, only include limited human representations, often too small or from unrecognizable angles (\cref{fig:face_recognition}(f)). Wang~\etal~\cite{wang2024lrdif} introduce still image datasets for facial expression recognition (FER), predicting emotions like fear, sadness, anger, and neutrality. However, these datasets are generated using a GAN trained on the P-OLED dataset, which does not adequately simulate actual UDC degradation, notably the lack of flare (\cref{fig:face_recognition}(e)). Additionally, these datasets are not publicly available. Conversely, UDC-VIT prominently features humans in 64.6\% of its videos (approved by the Institutional Review Board (IRB)), featuring various motions (\eg, hand waving, thumbs-up, body-swaying, and walking) by 22 carefully selected subjects from different angles (\cref{fig:face_recognition}(a)-(d)).

\vspace{-1mm}
\begin{figure}[!t]
\begin{minipage}{\linewidth}
   \centering
   \includegraphics[width=1.00\linewidth]{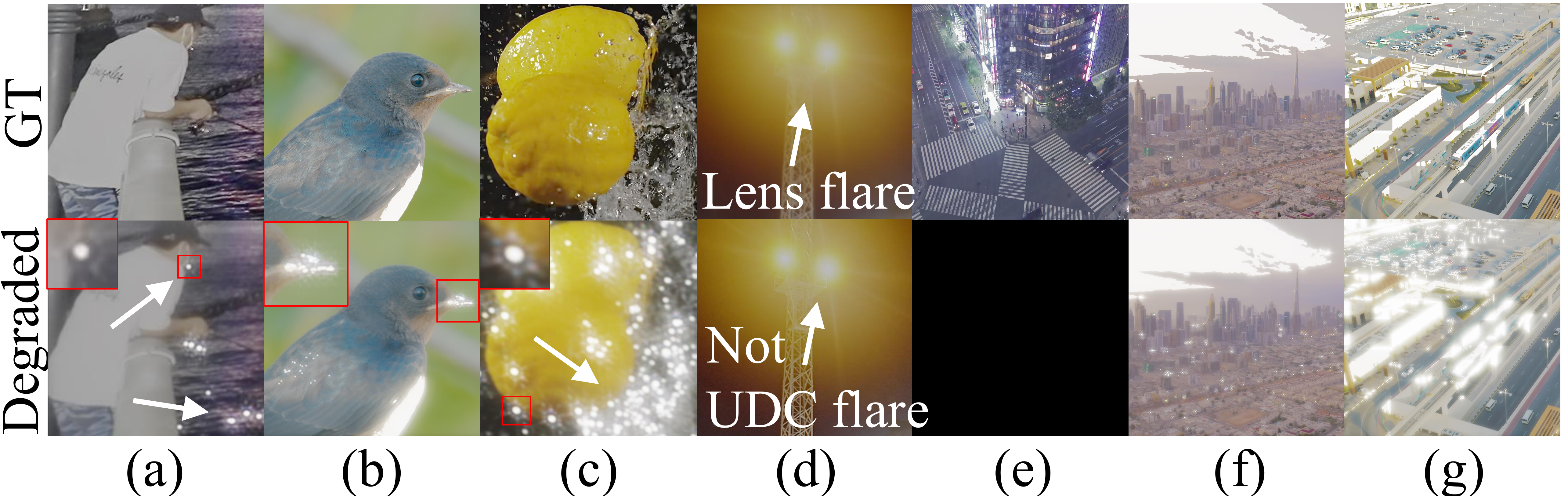}
   \caption{Strange scenes and white artifacts in VidUDC33K~\cite{liu2024decoupling}. (a) Flare in the seawater and on the cigarette. (b) Flare on the bird’s beak. (c) Flare on fruit. (d) Only lens flare is present, excluding UDC flare. (e) Black frame resulting from an incorrectly transformed PSF. (f) Washed-out colors on clouds. (g) Washed-out colors on buildings.}
\label{fig:less_meaningful}
\end{minipage}
\end{figure}


\paragraph{Strange scenes and white artifacts.}
The VidUDC33K dataset includes degraded frames with implausible flare artifacts, often appearing in unrelated regions, such as seawater, bird's beak, or fruits, as depicted in \cref{fig:less_meaningful}(a)-(c). These artifacts arise from flawed synthesis using PSF convolution after pixel scaling, which indiscriminately amplifies bright areas regardless of actual light sources. Some videos also lack UDC-specific flares altogether, instead presenting only natural lens flares or semantically empty scenes (\cref{fig:less_meaningful}(d)-(e)). In addition, excessive scaling introduces white artifacts that wash out natural colors in bright areas, such as clouds or white walls (\cref{fig:less_meaningful}(f)-(g)), raising concerns about the realism and practical utility of such data.  A theoretical analysis appears in Appendix~\ref{sec:vidudc33k_strange}.

\begin{table}[!t]
  \caption{The comparison of PCK\textsubscript{$\alpha$} values between the datasets. The UDC-VIT dataset showcases the best alignment quality. It has PCK values close to 100\% for all values of $\alpha$. ``\textit{Align Req.}" indicates whether an additional alignment process is required.}
  \label{tab:align_pck}
    \vspace{-0.9\baselineskip}
  \begin{center}
  \resizebox{\linewidth}{!}{%
  \begin{tabular}{l|c|c|cccc}
    \toprule
    \multirow{2}{*}{Dataset} & Align & Align      & \multirow{2}{*}{PCK\textsubscript{0.003}} & \multirow{2}{*}{PCK\textsubscript{0.01}} & \multirow{2}{*}{PCK\textsubscript{0.03}} & \multirow{2}{*}{PCK\textsubscript{0.10}} \\
                             & Req. & method &  &  &  &  \\ 
    \midrule
    T/P-OLED~\cite{zhou2021image}         &           &           & 96.32 & 98.11 & 98.45 & 99.08 \\
    SYNTH~\cite{feng2021removing}         &           &           & 99.70 & 99.95 & 99.96 & 99.99 \\
    Pseudo-real~\cite{feng2023generating} & \ding{52} & \small{AlignFormer}~\cite{feng2023generating} & N/A & \textbf{58.75} & \textbf{95.08} & \textbf{99.93} \\
    UDC-SIT~\cite{ahn2024udc}             & \ding{52} & DFT & \textbf{93.67} & \textbf{97.26} & \textbf{98.56} & \textbf{99.35} \\
    VidUDC33K~\cite{liu2024decoupling}    &           &           & 99.65 & 99.82 & 99.84 & 99.90 \\
    UDC-VIT                          & \ding{52} &           & \textbf{85.10} & \textbf{98.65} & \textbf{99.22} & \textbf{99.64} \\
    UDC-VIT                               & \ding{52} & DFT & \textbf{92.12} & \textbf{98.95} & \textbf{99.32} & \textbf{99.69} \\
    \bottomrule
  \end{tabular}
  }
  \end{center}
\end{table}

\paragraph{Alignment quality.}
\label{para:pck}
To assess the alignment quality of paired videos, we use LoFTR~\cite{sun2021loftr} as a keypoint matcher, following the convention of the previous studies~\cite{ahn2024udc,feng2023generating}. We compare the Percentage of Correct Keypoints (PCK), representing the ratio of correctly aligned keypoints to the total number. A keypoint pair is correctly aligned if $d < \alpha \times max(H, W)$, where $d$ is the positional difference between a pair of matched keypoints, $\alpha$ is the threshold, and $H$ and $W$ are the frame or image dimensions. We set $max(H, W) = 1024$ for fair comparison across datasets with varying resolutions.

\cref{tab:align_pck} compares alignment accuracy across datasets. The synthetic datasets (\eg, T-OLED/P-OLED, SYNTH, and VidUDC33K) do not require an additional alignment process, leading to PCK values near 100\%. In contrast, the Pseudo-real dataset using AlignFormer~\cite{feng2023generating}, attains a PCK value of 58.75\% for $\alpha = 0.01$. Unlike Pseudo-real, UDC-VIT achieves consistently high PCK values (92.12--99.69\%), demonstrating performance comparable to UDC-SIT, which previously led benchmarks.

\begin{table*}[!b]
  \caption{Restoration performance for synthetic and real UDC video datasets. The term \textit{Input} refers to the PSNR, SSIM, and LPIPS values between the degraded and ground-truth video pairs.} 
  \vspace{-0.9\baselineskip}
  \tiny
  \label{tab:basic_experiment}
  \begin{center}
  \resizebox{0.9\linewidth}{!}{
  \begin{tabular}{l|cc|ccc|ccc}
    \toprule
    & Runtime & Param & \multicolumn{3}{c|}{VidUDC33K~\cite{liu2024decoupling}} & \multicolumn{3}{c}{UDC-VIT} \\
    &  (sec)  &  (M)  & PSNR~$\uparrow$ & SSIM~$\uparrow$ & LPIPS~$\downarrow$ & PSNR~$\uparrow$ & SSIM~$\uparrow$ & LPIPS~$\downarrow$ \\
    \midrule
    Input                        &  -   &  -   & \textbf{26.22} & \textbf{0.8524} & \textbf{0.2642} & \textbf{16.26} & \textbf{0.7366} & \textbf{0.4117} \\
    DISCNet~\cite{feng2021removing}         & 0.73 & 3.80 & 28.89 & 0.8405 & 0.2432 & 24.70 & 0.8403 & 0.2675 \\
    UDC-UNet~\cite{liu2022udc}              & 0.37 & 5.70 & 28.37 & 0.8361 & 0.2561 & 28.00 & 0.8911 & 0.1779 \\
    FastDVDNet~\cite{tassano2020fastdvdnet} & 0.45 & 2.48 & 28.95 & 0.8638 & 0.2203 & 23.89 & 0.8439 & 0.2662 \\
    EDVR~\cite{wang2019edvr}                & 1.17 & 23.6 & 28.71 & 0.8531 & 0.2416 & 23.55 & 0.8331 & 0.2673 \\
    ESTRNN~\cite{zhong2020efficient}        & 0.20 & 2.47 & 29.54 & 0.8744 & 0.2170 & 25.38 & 0.8654 & 0.2216 \\
    DDRNet~\cite{liu2024decoupling}         & 0.44 & 5.76 & 31.91 & 0.9313 & 0.1306 & 24.68 & 0.8539 & 0.2218 \\
    \bottomrule
    \end{tabular}
    }
    \end{center}
\end{table*}

\section{Experiments}
\label{sec:experiments}
This section compares the UDC video restoration performance and face recognition accuracy of existing deep learning models trained by UDC-VIT and the previous synthetic video dataset. Please see Appendix~\ref{sec:apndx_cross} and~\ref{sec:analysis}, as well as \url{https://mcrl.github.io/UDC} for extensive analyses and visual comparisons, including cross-dataset validation, shooting conditions, and flickering artifacts in restored videos.

\subsection{Effects on Learnable Restoration Models}
In this section, we evaluate the effectiveness of the UDC-VIT dataset by comparing the video restoration performance of six deep learning models, each trained and tested on the UDC-VIT and VidUDC33K datasets~\cite{liu2024decoupling}. The comparison is performed only with VidUDC33K since PexelsUDC-T/P is not publicly available. DDRNet~\cite{liu2024decoupling} is the only existing UDC video restoration model, while FastDVDNet~\cite{tassano2020fastdvdnet}, EDVR~\cite{wang2019edvr}, and ESTRNN~\cite{zhong2020efficient} are video restoration models for other general tasks (\eg, deblur, denoising, and super-resolution). DISCNet~\cite{feng2021removing} and UDC-UNet~\cite{liu2022udc} are UDC still image restoration models.

\cref{tab:basic_experiment} shows the restoration performance of the six models on both VidUDC33K and UDC-VIT. Interestingly, the performance rankings of the benchmark models across the two datasets do not consistently align. The varying severity of flares between the two datasets is the main reason for the inconsistent restoration performance rankings. Unlike UDC-VIT, VidUDC33K does not accurately represent real-world degradations (e.g., flares), as reflected in its higher input-level PSNR and SSIM, and lower LPIPS. The top-performing models on UDC-VIT, such as UDC-UNet and ESTRNN, use residual CNNs to address complex degradations and improve restoration quality. They also provide better frame-to-frame consistency than the others, which is crucial for reducing flicker, although some flicker persists. This shows the benefits of residual connections in improving consistency. Note that the restored video of VidUDC33K by DDRNet, using their pre-trained model, does not exhibit flickering. This highlights the need for real-world UDC video datasets, such as UDC-VIT, and is also evident in cross-dataset experiments presented in Appendix~\ref{sec:apndx_cross}, which demonstrate the limitations of synthetic data in modeling complex degradations.

\subsection{Face Recognition}
\begin{figure*}[!t]
  \centering
  \begin{minipage}{\linewidth}
    \centering
    \includegraphics[width=0.7\linewidth]{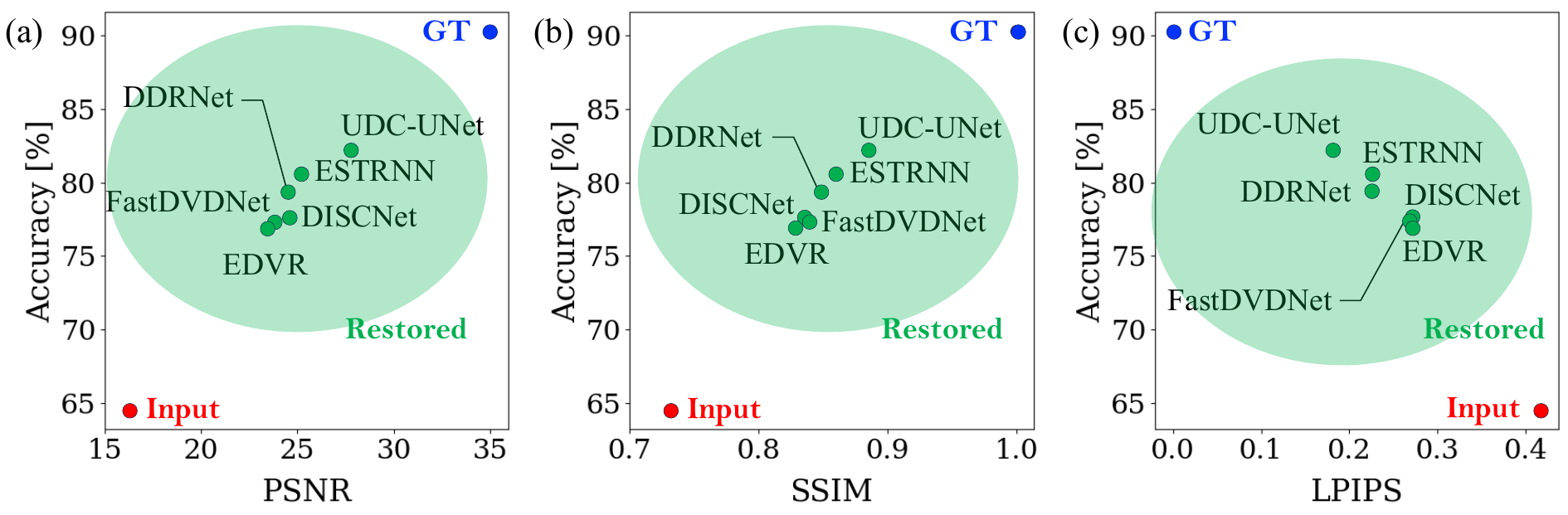}
    \vspace{-0.3\baselineskip}
    \caption{FR accuracy. Model error cases are excluded when calculating the accuracy, where the model error indicates when FR models fail due to severe UDC degradation. Frames restored by deep learning models with higher performance in (a) PSNR, (b) SSIM, and (c) LPIPS achieve better FR accuracy. PSNR between the two GTs (i.e., \texttt{\textcolor{blue}{GT}}) is plotted as 35.00 for easy observation.}
    \label{fig:experiment_face}
  \end{minipage}
\end{figure*}

The face recognition (FR) task verifies whether two images are of the same person, similar to typical smartphone applications like Face ID. As shown in \cref{fig:experiment_face}, we assess average FR accuracy using seven FR models from the DeepFace library~\cite{deepface}, such as VGG-Face~\cite{parkhi2015deep}, Facenet~\cite{schroff2015facenet}, OpenFace~\cite{baltruvsaitis2016openface}, DeepFace~\cite{taigman2014deepface}, DeepID~\cite{sun2014deep}, Dlib~\cite{king2009dlib}, and ArcFace~\cite{deng2019arcface}. We test 600 FR frame pairs (human 1 and human 2 \textit{from different videos}) on a balanced dataset, with 49.2\% of the same person (human 1 = human 2) and 50.8\% of different people (human 1 $\not =$  human 2). Note that \cref{tab:basic_experiment} ranks models on the complete UDC-VIT test set, while \cref{fig:experiment_face} evaluates only a sampled subset of the face-related data. 

As shown in \cref{fig:experiment_face}, we compare the effect of human 2's restoration level in terms of PSNR, SSIM, and LPIPS ($X$-axis) on FR accuracy ($Y$-axis). When evaluating FR accuracy, human 1 is always ground truth (GT), and human 2 can be \texttt{\textcolor{red}{Input}}, \texttt{\textcolor{mygreen}{Restored}}, or \texttt{\textcolor{blue}{GT}}. Therefore, \texttt{\textcolor{red}{Input}}, \texttt{\textcolor{mygreen}{Restored}}, or \texttt{\textcolor{blue}{GT}} in \cref{fig:experiment_face} indicates the group to which human 2 belongs. For example, in \cref{fig:experiment_face}(a), \texttt{\textcolor{red}{Input}}'s PSNR is calculated between human 2 (\texttt{\textcolor{red}{Input}}) and human 2 (GT). Similarly, \texttt{\textcolor{red}{Input}}'s FR accuracy is calculated between human 1 (GT) and human 2 (\texttt{\textcolor{red}{Input}}). To verify the relationship between restoration level and FR accuracy, we illustrate six deep-learning models' restoration performance (highlighted with a green circle) and corresponding FR accuracy in \cref{fig:experiment_face}. \texttt{\textcolor{mygreen}{Restored}}'s PSNR is calculated between human 2 (\texttt{\textcolor{mygreen}{Restored}}) and human 2 (GT). Similarly, \texttt{\textcolor{mygreen}{Restored}}'s FR accuracy is calculated between human 1 (GT) and human 2 (\texttt{\textcolor{mygreen}{Restored}}).

The results demonstrate the importance of leveraging UDC degradation by deep-learning restoration models to improve FR accuracy, consistent with prior work~\cite{ahn2025investigating}. For example, as depicted in \cref{fig:experiment_face}(a), \texttt{\textcolor{red}{Input}} with PSNR of 16.31 shows 64.5\% FR accuracy, \texttt{\textcolor{mygreen}{UDC-UNet}} with PSNR of 27.74 shows 82.2\% FR accuracy, and \texttt{\textcolor{blue}{GT}} shows 90.3\% FR accuracy.

\section{Limitations}
\label{sec:limitations}
UDC-VIT has three main limitations. First, UDC degradations vary with display pixel design, affecting diffraction patterns, PSF, and light propagation. As a result, models trained on UDC-VIT may not work optimally on devices other than Samsung Galaxy Z-Fold 5~\cite{samsungGalaxyZFold5}, such as the ZTE Axon series~\cite{zteAxon20,zteAxon30,zteAxon40-ultra} or other Samsung Galaxy Z-Fold series~\cite{samsungGalaxyZFold3,samsungGalaxyZFold4,samsungGalaxyZFold6}. However, transfer learning can help adapt models to new devices (Please see Appendix~\ref{sec:apndx_cross}). Second, fast-moving objects, such as speeding cars, are excluded due to potential frame-to-frame discrepancies, even with tight synchronization ($<$ 8 msec). Third, UDC restoration is inherently hardware-aware, as degradations depend on the designs of the optics, sensor, and panel. This underscores the importance of hardware–software co-design and motivates research into generalizable software solutions that can be applied across various devices.

\section{Conclusion}
\label{sec:conclusion}
To the best of our knowledge, UDC-VIT is the first UDC video dataset that includes actual UDC degradation, such as low transmittance, blur, noise, and flare. We propose an efficient video-capturing system to acquire matched pairs of UDC-degraded and ground-truth videos, ensuring precise synchronization between the two cameras. We further align the UDC-VIT frame by frame using the DFT, achieving sufficient alignment accuracy to train deep learning models. We demonstrate the effectiveness of UDC-VIT by comparing it with other datasets. Notably, UDC-VIT solely presents significant actual UDC degradation (\eg, \textit{variant flares}) and stands out from other datasets by featuring videos tailored for face recognition. Our experiments demonstrate that models trained on synthetic UDC videos are impractical, as they fail to capture the actual degradation characteristics accurately. Moreover, we show that restoring UDC degradation significantly improves face recognition accuracy. The UDC-VIT dataset can be downloaded from \url{https://github.com/mcrl/UDC-VIT}.

\section*{Acknowledgment}
This work was partially supported by the National Research Foundation of Korea (NRF) under Grant No. RS-2023-00222663 (Center for Optimizing Hyperscale AI Models and Platforms), and by the Institute for Information and Communications Technology Promotion (IITP) under Grant No. 2018-0-00581 (CUDA Programming Environment for FPGA Clusters) and No. RS-2025-02304554 (Efficient and Scalable Framework for AI Heterogeneous Cluster Systems), all funded by the Ministry of Science and ICT (MSIT) of Korea. Additional support was provided by the BK21 Plus Program for Innovative Data Science Talent Education (Department of Data Science, SNU, No. 5199990914569) and the BK21 FOUR Program for Intelligent Computing (Department of Computer Science and Engineering, SNU, No. 4199990214639), both funded by the Ministry of Education (MOE) of Korea. This work was also partially supported by the Artificial Intelligence Industrial Convergence Cluster Development Project, funded by the MSIT and Gwangju Metropolitan City. It was also supported in part by Samsung Display Co., Ltd. Research facilities were provided by ICT at Seoul National University.

{
    \small
    \bibliographystyle{ieeenat_fullname}
    \bibliography{main}
}

\appendix

\setcounter{figure}{0}  
\setcounter{table}{0} 
\setcounter{algorithm}{0} 
\counterwithin{figure}{section}
\counterwithin{table}{section}
\counterwithin{algorithm}{section}

\clearpage
\maketitlesupplementary

\section{Analyzing the Limitations of Synthetic Datasets}
\label{sec:vidudc33k_strange}

\begin{figure*}[!b]
   \centering
   \vspace{8.0mm}
   \includegraphics[width=1.0\linewidth]{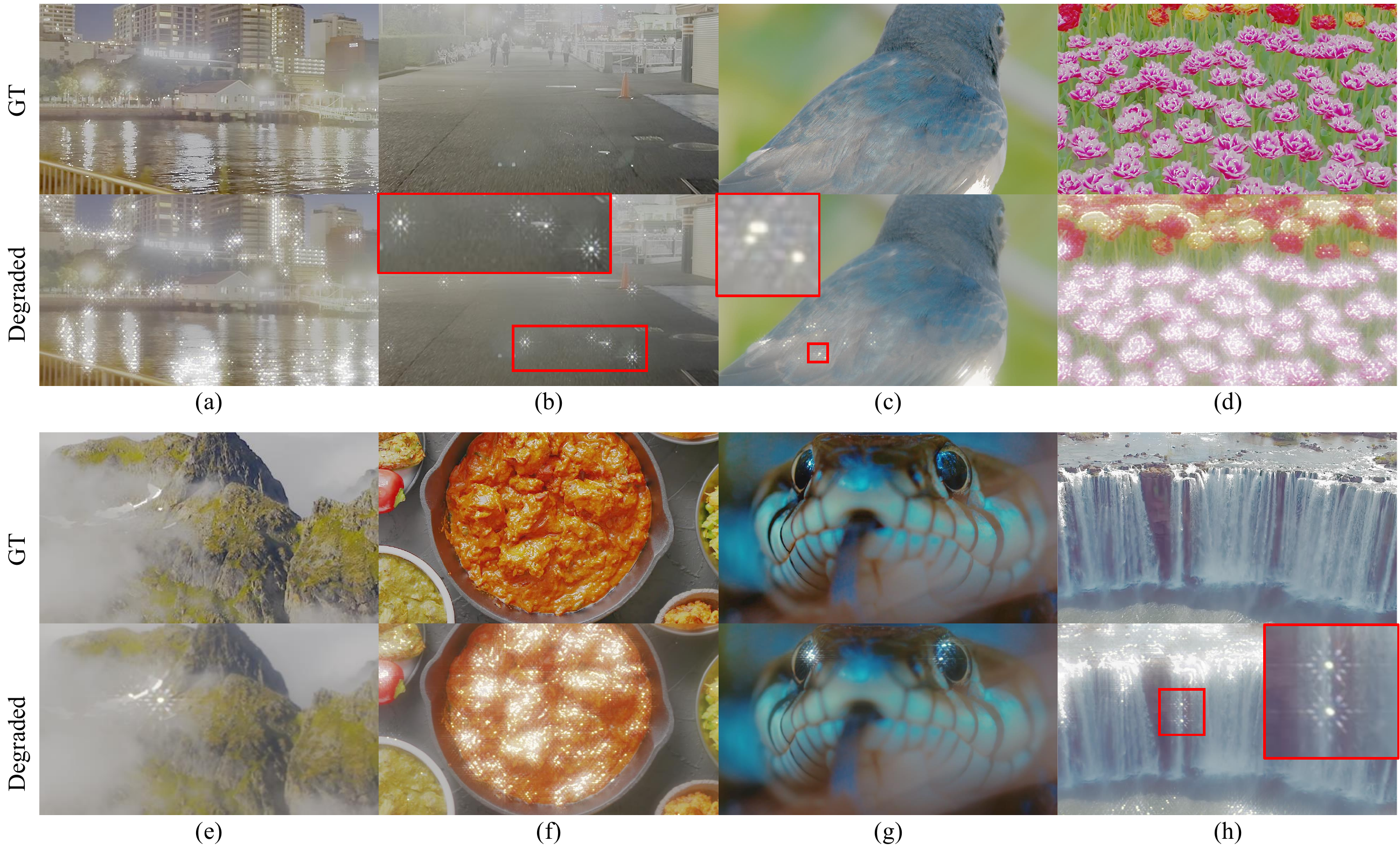}
   \vspace{0.3mm}
   \caption{The visual illustration that showcases improbable flares resulting from excessive scaling in the VidUDC33K dataset~\cite{liu2024decoupling}. (a) Flare in the river. (b) Flare from the dust on the camera lens. (c) Flare on the bird feathers. (d) Flare on the flower petals. (e) Flare on the mountain peaks. (f) Flare on the food. (g) Flare in the snake eyes. (h) Flare on the waterfalls.
}
\label{fig:improbable_flares}
\end{figure*}

This section provides a comprehensive analysis of the limitations of synthetic datasets (\eg, VidUDC33K~\cite{liu2024decoupling}). As outlined in \cref{sec:datasets_comparison} in the main body, the VidUDC33K dataset exhibits several strange scenes. In this section, we analyze three representative anomalies frequently observed in the dataset: \textit{flares occurring in physically improbable scenarios}, \textit{unintended white artifacts}, and \textit{darkened and nearly featureless degraded frames}.

\paragraph{Improbable situations and unintended white artifacts.}
Liu~\etal~\cite{liu2024decoupling} attempt to synthesize flares by convolving the PSF with ground-truth images. However, the desired flares do not manifest as expected. To address this, they apply a scaling procedure to pixel values exceeding a predefined threshold to amplify these intensities, followed by PSF convolution. 

\vspace{3.0mm}

This process leads to two notable phenomena. First, \textit{flares appear in physically improbable scenarios}. Their method generates flares with values of white pixels exceeding the threshold regardless of whether these correspond to actual light sources, as shown in \cref{fig:less_meaningful}(a)-(c) in the main body and \cref{fig:improbable_flares}. 

\begin{figure*}[t]
   \centering
   \includegraphics[width=0.95\linewidth]{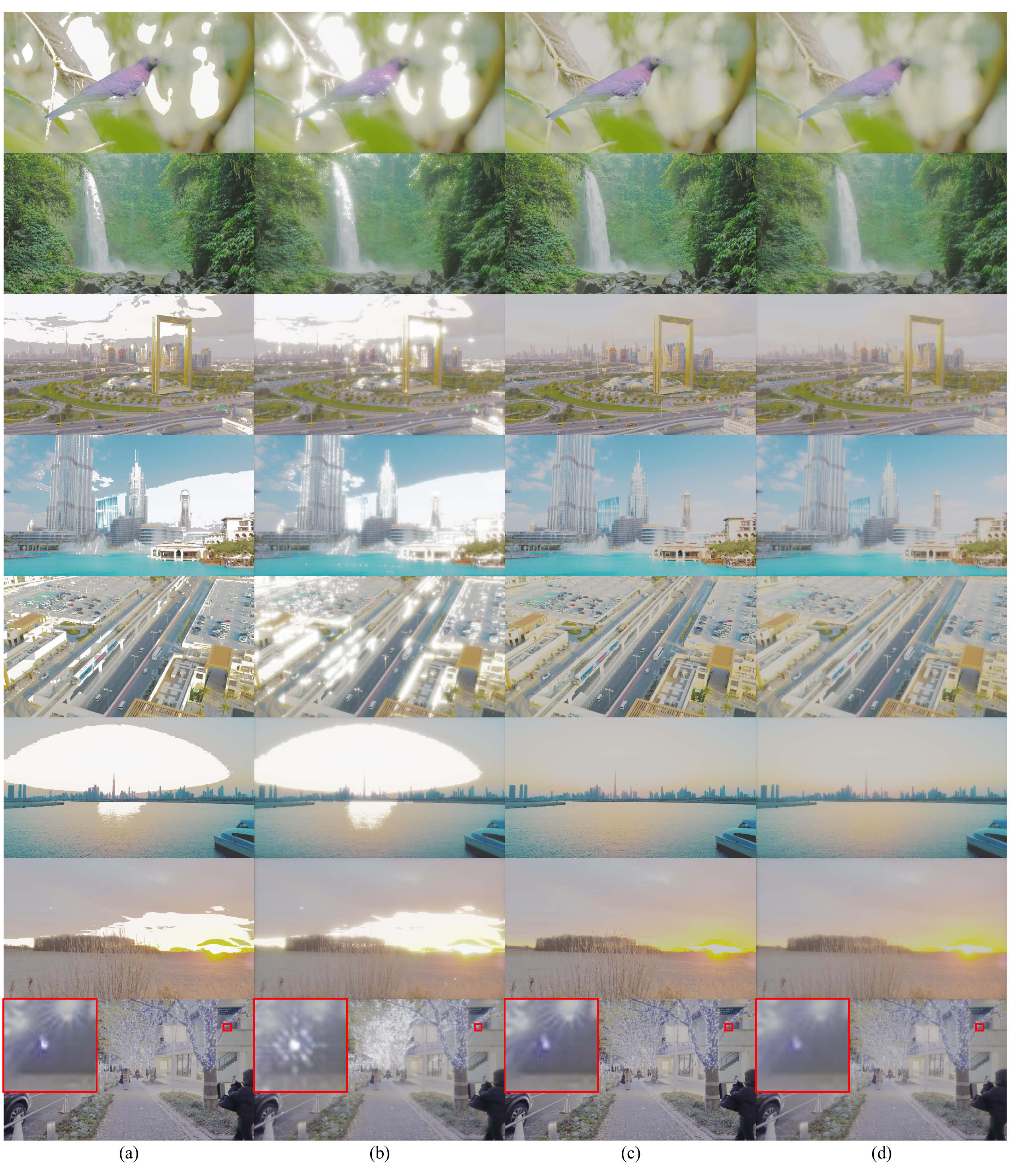}
   \caption{The visual depiction that shows white artifacts resulting from excessive scaling in the VidUDC33K dataset~\cite{liu2024decoupling}. The frames processed \textit{without} the scaling procedure do not exhibit these artifacts (see (c) and (d)), in contrast to the frames processed \textit{with} the scaling procedure (see (a) and (b)). Notably, flares are visible only when the scaling procedure is applied (see (b)), while absent without it (see (d)). This suggests that the authors rely on scaling to generate flares, unintentionally producing unrealistic white artifacts. 
   (a) The ground-truth frame \textit{with} scaling procedure.
   (b) The degraded frame \textit{with} scaling procedure.
   (c) The ground-truth frame \textit{without} scaling procedure.
   (d) The degraded frame \textit{without} scaling procedure.
}
\label{fig:white_artifacts}
\end{figure*}

Second, \textit{unintended white artifacts} occur. Since pixel values above a predefined threshold are amplified, some regions become saturated, appearing as white, and lose their original colors. For example, areas with clouds in the sky, waterfalls, and white walls become excessively white, as depicted in \cref{fig:less_meaningful}(f)-(g) in the main body and \cref{fig:white_artifacts}. To verify the relationship between the scaling and flare generation, we conducted an experiment shown in the bottom row of \cref{fig:white_artifacts}. Without the scaling procedure, flares fail to appear even in frames where they are expected, as illustrated in the bottom row of \cref{fig:white_artifacts}(d). Approximately 12\% of the videos exhibit these unintended white artifacts caused by the scaling, which makes the data unsuitable for deep-learning training.

\begin{figure*}[!t]
   \centering
   \includegraphics[width=0.99\linewidth]{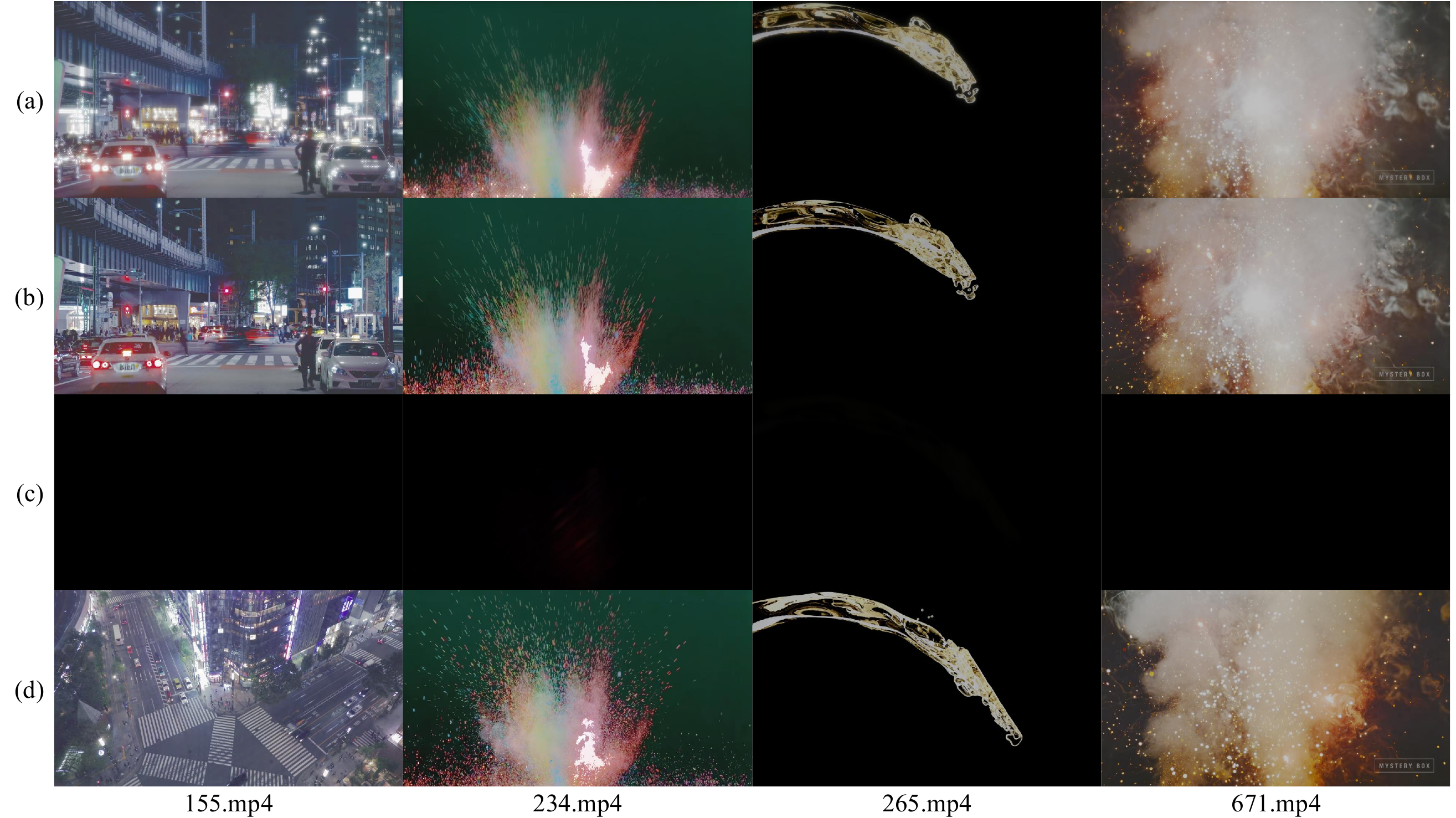}
   \caption{The visual representation that demonstrates black frames resulting from incorrectly transformed PSFs in the VidUDC33K dataset~\cite{liu2024decoupling}.
   (a) The first frame of the degraded video.
   (b) The first frame of the ground-truth video.
   (c) The tenth frame of the degraded video.
   (d) The tenth frame of the ground-truth video.
   }
\label{fig:black_frame}
\end{figure*}

\paragraph{The darkened and nearly featureless frames.}
Liu~\etal~\cite{liu2024decoupling} strive to create temporally variant flares in continuous video sequences. They simulate the dynamic changes of the PSF during motion by computing the inter-frame homography matrix $H_{t-1 \rightarrow t}$, formulated as \cref{eq:homography}, between consecutive frames. 

\begin{equation}
  \label{eq:homography}
  \begin{aligned}
    k_t &= \mathcal{T}(k_{t-1}, H_{t-1 \rightarrow t}) \\
    &= \bigg| \mathcal{F} \left( H^{-1}_{t-1 \rightarrow t} \left( \mathcal{F}^{-1} \left( \sqrt{k_{t-1}} \right) \right) \right) \bigg| ^2, \\
    H_{t-1 \rightarrow t} &= \mathcal{M}(I_{t-1}^{GT}, I_t^{GT}),
  \end{aligned}
\end{equation}

where $\mathcal{T}(\cdot)$ is the transformation function that utilizes $H^{-1}_{t-1 \rightarrow t}$ to perform a perspective warp on the PSF of the previous frame, $k_{t-1}$. $H^{-1}_{t-1 \rightarrow t}$ denotes the inverse matrix of $H_{t-1 \rightarrow t}$. $\mathcal{F}(\cdot)$ and $\mathcal{F}^{-1}(\cdot)$ represent the Fourier transform and its inverse, respectively. $\mathcal{M}(\cdot)$ is the matching component used to calculate the homography matrix between frames.

However, this process occasionally results in PSF values approaching zero, causing the degraded frames to appear entirely black. Specifically, this issue occurs in 4 out of 677 videos. The first frame does not undergo PSF transformation, while subsequent frames do. Therefore, as seen in \cref{fig:less_meaningful}(e) in the main body and \cref{fig:black_frame}(c), the frames after the first one (\eg, the tenth frame) sometimes become black.

\section{Cross-dataset Validation}
\label{sec:apndx_cross}

\begin{figure*}[t!]
\begin{minipage}{\linewidth}
   \centering
   \includegraphics[width=0.7\linewidth]{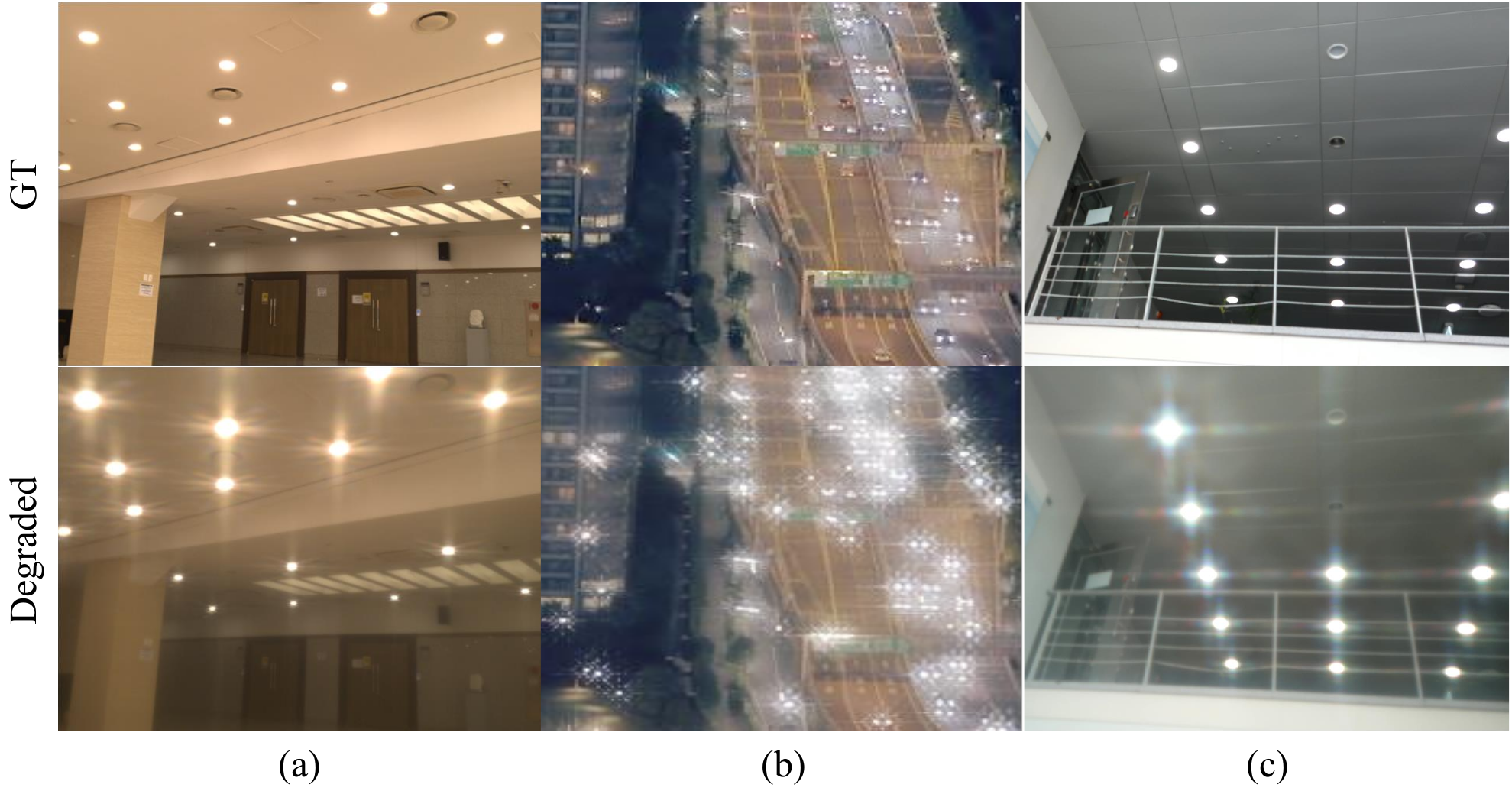}
   \caption{Comparison of the UDC datasets, showing varied data distribution and degradation patterns. (a) UDC-SIT~\cite{ahn2024udc}. (b) VidUDC33K~\cite{liu2024decoupling}. (c) UDC-VIT.}
\label{fig:apndx_cross_intro}
\end{minipage}
\end{figure*}

This section demonstrates cross-dataset validation to address the unique dataset distribution and degradation patterns of UDC datasets, as discussed in \cref{sec:limitations}. Transfer learning techniques are crucial for addressing varying dataset distributions and degradation patterns in practical UDC restoration. These methods include approaches such as fine-tuning and domain adaptation~\cite{gou2024test,kim2020transfer,du2020learning}). In this section, we focus on fine-tuning and evaluate its effectiveness across multiple UDC datasets.

To better illustrate the challenges in cross-dataset generalization and motivate the need for transfer learning, we analyze device-level differences across representative UDC datasets. For example, Samsung Galaxy Z-Fold 5 (UDC-VIT) and ZTE Axon 20~\cite{zteAxon20} (VidUDC33K~\cite{liu2024decoupling}) have vastly different pixel designs, as they come from different vendors. Similarly, Samsung Galaxy Z-Fold 3~\cite{samsungGalaxyZFold3} (UDC-SIT~\cite{ahn2024udc}) and Samsung Galaxy Z-Fold 5~\cite{samsungGalaxyZFold5} (UDC-VIT) share relatively similar designs, but they still exhibit differences.

\begin{table*}[!t]
  \caption{The design of experiments to verify the effect of fine-tuning and the use of a real-world dataset (\eg, UDC-VIT). The first and the second subscripts beside $\mathcal{M}$ indicate the training and fine-tuning datasets, respectively. For example, $\mathcal{M}_{s3}$ refers to the model trained on UDC-SIT without fine-tuning, while $\mathcal{M}_{s3s5}$ denotes the model trained on UDC-SIT and subsequently fine-tuned on UDC-VIT. Models without subscripts are trained and tested on the same dataset.}
  \label{tab:model_finetuning}
  \begin{center}
  \resizebox{0.65\linewidth}{!}{
  \begin{tabular}{ll|ccc}
    \toprule
    Experiments & Model name & Training dataset & Fine-tuning dataset & Test dataset \\ 
    \midrule
    \multirow{3}{*}{Exp. 1} & $\mathcal{M}_{s3}$    & UDC-SIT   &    -      & UDC-VIT   \\
                            & $\mathcal{M}_{s3s5}$  & UDC-SIT   & UDC-VIT   & UDC-VIT   \\ 
                            & $\mathcal{M}$         & UDC-VIT   &    -      & UDC-VIT   \\ \cline{1-5}
    \multirow{3}{*}{Exp. 2} & $\mathcal{M}_{s5}$    & UDC-VIT   &    -      & VidUDC33K \\
                            & $\mathcal{M}_{s5z20}$ & UDC-VIT   & VidUDC33K & VidUDC33K \\ 
                            & $\mathcal{M}$         & VidUDC33K &    -      & VidUDC33K \\ \cline{1-5}
    \multirow{3}{*}{Exp. 3} & $\mathcal{M}_{z20}$   & VidUDC33K &    -      & UDC-VIT   \\
                            & $\mathcal{M}_{z20s5}$ & VidUDC33K & UDC-VIT   & UDC-VIT   \\    
                            & $\mathcal{M}$         & UDC-VIT   &    -      & UDC-VIT   \\
    \bottomrule
  \end{tabular}
  }
  \end{center}
\end{table*}

\vspace{\baselineskip}

\cref{fig:apndx_cross_intro}(a) and (c) illustrate that the UDC-SIT and UDC-VIT datasets show similar degradation, such as blur, decrease in transmittance, and flare shape. In contrast, \cref{fig:apndx_cross_intro}(b) and (c) highlight the stark difference between the VidUDC33K and UDC-VIT datasets. This discrepancy arises from two factors: the variation in pixel design and the synthetic nature of the VidUDC33K dataset, which results in unrealistic degradation patterns.

Fine-tuning models to address variant dataset distributions or degradation patterns is crucial in practical applications. To evaluate the effect of fine-tuning and validate the effectiveness of UDC-VIT, which reflects real-world degradation, we conduct three experiments (Exp. 1-3), as shown in \cref{tab:model_finetuning}. The subscripts beside the model name $\mathcal{M}$ specify the datasets used for training and fine-tuning. For example, $\mathcal{M}_{s3}$ refers to the model trained on UDC-SIT (\textbf{S}amsung Galaxy Z-Fold \textbf{3}) without fine-tuning, $\mathcal{M}_{s5z20}$ is trained on UDC-VIT  (\textbf{S}amsung Galaxy Z-Fold \textbf{5}) and fine-tuned on VidUDC33K (\textbf{Z}TE Axon \textbf{20}), while $\mathcal{M}_{z20s5}$ is trained on VidUDC33K (\textbf{Z}TE Axon \textbf{20}) and fine-tuned on UDC-VIT (or \textbf{S}amsung Galaxy Z-Fold \textbf{5}). We use models $\mathcal{M}$ such as UDC-UNet~\cite{liu2022udc}, DISCNet~\cite{feng2021removing}, and DDRNet~\cite{liu2024decoupling} among six benchmark models in \cref{tab:basic_experiment}. Fine-tuning is performed for 10\% or 20\% of the total iterations, with the learning rate set to 10\% or 20\% of the original value.

\paragraph{Experiment 1: impact of fine-tuning on UDC-VIT.}

\begin{table*}[b]
  \caption{\textbf{[Exp. 1]} Restoration performance of DISCNet~\cite{feng2021removing} and UDC-UNet~\cite{liu2022udc} when tested on UDC-VIT. They are trained on UDC-SIT~\cite{ahn2024udc}, either with or without fine-tuning on UDC-VIT, or solely trained on UDC-VIT. Models without subscripts refer to those solely trained on UDC-VIT, meaning their PSNR, SSIM, and LPIPS values match those in  \cref{tab:basic_experiment} in the main body. The number of iterations represents the percentage of fine-tuning iterations relative to the total iterations in the original configurations provided by the authors.}
  \label{tab:apndx_cross_exp1}
  \begin{center}
  \resizebox{0.78\linewidth}{!}{%
  \begin{tabular}{l|cccccc}
    \toprule
    Model name & PSNR~$\uparrow$ & SSIM~$\uparrow$ & LPIPS~$\downarrow$ & Training & Fine-tuning ($\# Iterations$) & Test \\ 
    \midrule
    $\text{DISCNet}_{s3}$    & 16.81 & 0.7139 & 0.3293 & $\text{UDC-SIT}$ &       -        & $\text{UDC-VIT}$ \\
    $\text{DISCNet}_{s3s5}$  & 23.16 & 0.8281 & 0.2527 & $\text{UDC-SIT}$ & $\text{UDC-VIT}$ (10\%) & $\text{UDC-VIT}$ \\
    $\text{DISCNet}_{s3s5}$  & 23.57 & 0.8331 & 0.2459 & $\text{UDC-SIT}$ & $\text{UDC-VIT}$ (20\%) & $\text{UDC-VIT}$ \\ 
    $\text{DISCNet}$         & 24.70 & 0.8403 & 0.2675 & $\text{UDC-VIT}$ &       -        & $\text{UDC-VIT}$ \\ \cline{1-7}
    $\text{UDC-UNet}_{s3}$   & 17.21 & 0.7260 & 0.3400 & $\text{UDC-SIT}$ &       -        & $\text{UDC-VIT}$ \\
    $\text{UDC-UNet}_{s3s5}$ & 24.94 & 0.8709 & 0.2113 & $\text{UDC-SIT}$ & $\text{UDC-VIT}$ ($10\%$) & $\text{UDC-VIT}$ \\
    $\text{UDC-UNet}_{s3s5}$ & 25.41 & 0.8758 & 0.2015 & $\text{UDC-SIT}$ & $\text{UDC-VIT}$ ($20\%$) & $\text{UDC-VIT}$ \\
    $\text{UDC-UNet}$        & 28.00 & 0.8911 & 0.1779 & $\text{UDC-VIT}$ &       -        & $\text{UDC-VIT}$ \\
    \bottomrule
  \end{tabular}
  }
  \end{center}
\end{table*}

\begin{figure*}[!b]
   \centering
   \includegraphics[width=0.95\linewidth]{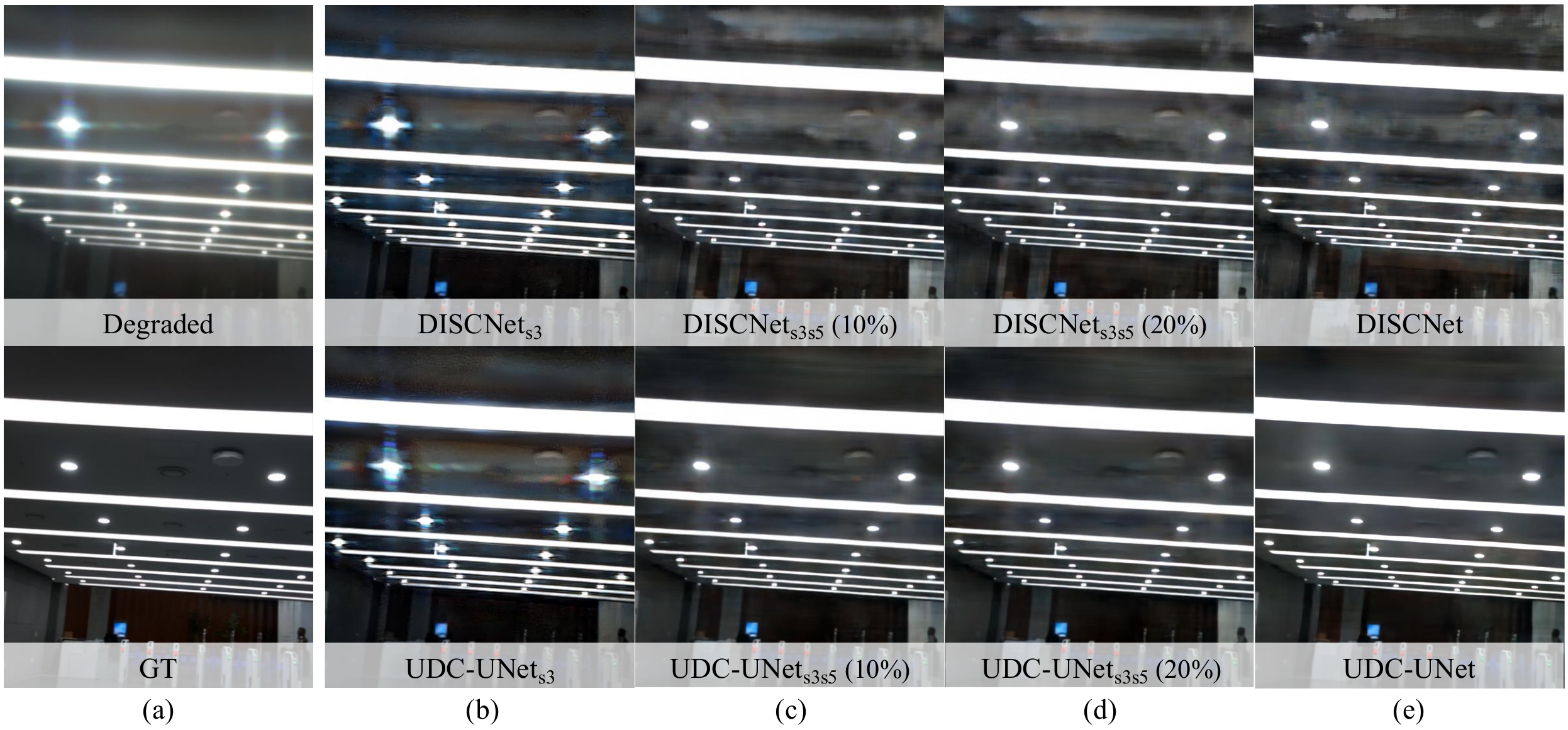}
   \caption{\textbf{[Exp. 1]} Comparison of restoration performance across different models when tested on the UDC-VIT dataset. (a) UDC-degraded (at the first row) and GT (at the second row) frames of the UDC-VIT dataset. Test frames from UDC-VIT by DISCNet (at the first row) and UDC-UNet (at the second row) trained on UDC-SIT (b) without fine-tuning on UDC-VIT, (c) with fine-tuning on UDC-VIT (10\%), and (d) with fine-tuning on UDC-VIT (20\%). (e) Test frames from UDC-VIT by DISCNet (at the first row) and UDC-UNet (at the second row), solely trained on UDC-VIT.}
\label{fig:apndx_cross_exp1}
\end{figure*}

This experiment evaluates the impact of fine-tuning on UDC-VIT when tested on UDC-VIT. We compare the performance of models trained on UDC-SIT, with or without fine-tuning on UDC-VIT, and models trained solely on UDC-VIT. Since UDC-SIT is a still image dataset, we use UDC-UNet and DISCNet as models $\mathcal{M}$, which are specifically designed for UDC still images. As presented in \cref{tab:apndx_cross_exp1}, $\text{DISCNet}_{s3}$ and $\text{UDC-UNet}_{s3}$ trained exclusively on UDC-SIT struggle to generalize to UDC-VIT. In contrast, $\text{DISCNet}_{s3s5}$ and $\text{UDC-UNet}_{s3s5}$, which incorporate fine-tuning with UDC-VIT, demonstrate superior restoration performance for UDC-VIT degradations. In particular, increasing the number of fine-tuning iterations further enhances the performance. This is also evident in \cref{fig:apndx_cross_exp1}. Models without fine-tuning (\eg, $\text{DISCNet}_{s3}$ and $\text{UDC-UNet}_{s3}$) struggle to restore UDC-VIT, as shown in \cref{fig:apndx_cross_exp1}(b). In contrast, models with fine-tuning (\eg, $\text{DISCNet}_{s3s5}$ and $\text{UDC-UNet}_{s3s5}$), as shown in \cref{fig:apndx_cross_exp1}(c) and (d), achieve restoration performance comparable to those solely trained on UDC-VIT (\eg, $\text{DISCNet}$ and $\text{UDC-UNet}$), as depicted in \cref{fig:apndx_cross_exp1}(e).

These findings lead to the following conclusions. Although the Samsung Galaxy Z-Fold 3 (UDC-SIT) and Samsung Galaxy Z-Fold 5 (UDC-VIT) share similar pixel designs, being from the same vendor, their differences are substantial enough to require fine-tuning. With adequate adaptation, however, these models effectively leverage degradations from other UDC devices, underscoring the potential for cross-device generalization with fine-tuning.

\paragraph{Experiment 2: impact of fine-tuning on VidUDC33K.}

This experiment aims to assess the impact of fine-tuning on VidUDC33K when tested on VidUDC33K. It compares the performance of models trained on UDC-VIT, with or without fine-tuning on VidUDC33K, and models trained solely on VidUDC33K. For models $\mathcal{M}$, we use UDC-UNet, DISCNet, and DDRNet, which are explicitly designed to address UDC degradations. 

\begin{table*}[!b]
  \caption{\textbf{[Exp. 2]} Restoration performance of DISCNet~\cite{feng2021removing}, UDC-UNet~\cite{liu2022udc}, and DDRNet~\cite{liu2024decoupling} when tested on VidUDC33K~\cite{liu2024decoupling}. They are trained on UDC-VIT, either with or without fine-tuning on VidUDC33K, or solely trained on VidUDC33K. Models without subscripts refer to those solely trained on VidUDC33K, meaning their PSNR, SSIM, and LPIPS values match those in \cref{tab:basic_experiment} in the main body. The number of iterations represents the percentage of fine-tuning iterations relative to the total iterations in the original configurations provided by the authors.}
  \label{tab:apndx_cross_exp2}
  \begin{center}
  \resizebox{0.8\linewidth}{!}{%
  \begin{tabular}{l|cccccc}
    \toprule
    Model name & PSNR~$\uparrow$ & SSIM~$\uparrow$ & LPIPS~$\downarrow$ & Training & Fine-tuning ($\# Iterations$) & Test \\ 
    \midrule
    $\text{DISCNet}_{s5}$     & 18.73 & 0.7503 & 0.4159 & UDC-VIT   &         -        & VidUDC33K \\
    $\text{DISCNet}_{s5z20}$  & 28.89 & 0.9129 & 0.1727 & UDC-VIT   & VidUDC33K ($10\%$) & VidUDC33K \\
    $\text{DISCNet}$          & 28.89 & 0.8405 & 0.2432 & VidUDC33K &         -        & VidUDC33K \\ \cline{1-7}
    $\text{UDC-UNet}_{s5}$    & 19.84 & 0.7682 & 0.3737 & UDC-VIT   &         -        & VidUDC33K \\
    $\text{UDC-UNet}_{s5z20}$ & 29.57 & 0.9139 & 0.1506 & UDC-VIT   & VidUDC33K ($10\%$) & VidUDC33K \\
    $\text{UDC-UNet}$         & 28.37 & 0.8361 & 0.2561 & VidUDC33K &         -        & VidUDC33K \\ \cline{1-7}
    $\text{DDRNet}_{s5}$      & 20.10 & 0.8313 & 0.3446 & UDC-VIT   &         -        & VidUDC33K \\
    $\text{DDRNet}_{s5z20}$   & 29.12 & 0.8994 & 0.2073 & UDC-VIT   & VidUDC33K ($10\%$) & VidUDC33K \\
    $\text{DDRNet}$           & 31.91 & 0.9313 & 0.1306 & VidUDC33K &         -        & VidUDC33K \\
    \bottomrule
  \end{tabular}
  }
  \end{center}
\end{table*}

\begin{figure*}[!b]
   \centering
   \includegraphics[width=0.9\linewidth]{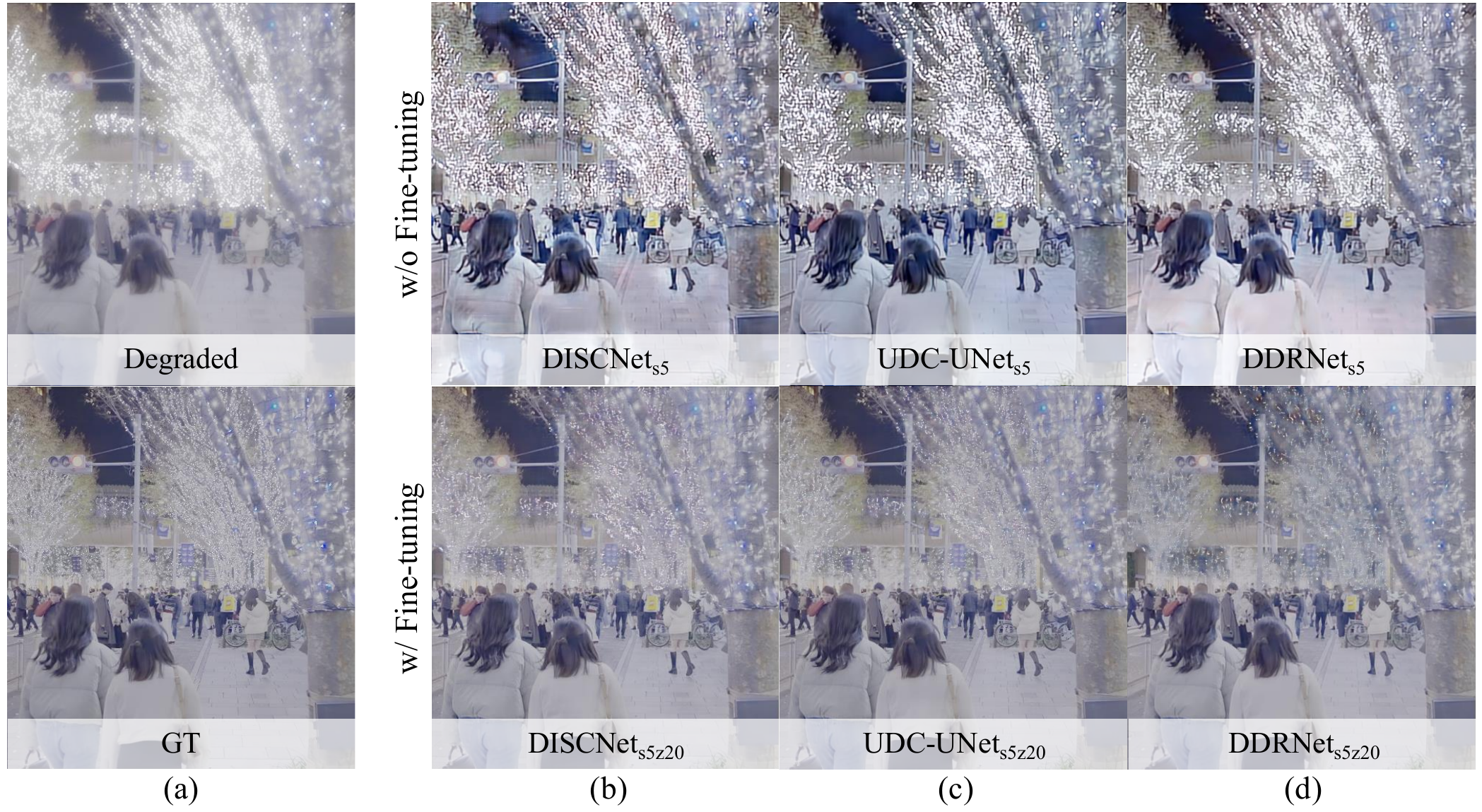}
   \caption{\textbf{[Exp. 2]} Comparison of restoration performance across different models when tested on the VidUDC33K dataset~\cite{liu2024decoupling}. (a) UDC-degraded (at the first row) and GT (at the second row) frames of the VidUDC33K dataset. The models in the first row are pre-trained on UDC-VIT without fine-tuning on VidUDC33K: (b) $\text{DISCNet}_{s5}$. (c) $\text{UDC-UNet}_{s5}$. (d) $\text{DDRNet}_{s5}$. The models in the second row are pre-trained on UDC-VIT with fine-tuning on VidUDC33K: (b) $\text{DISCNet}_{s5z20}$. (c) $\text{UDC-UNet}_{s5z20}$. (d) $\text{DDRNet}_{s5z20}$, showing improved restoration performance.}
\label{fig:apndx_cross_exp2}
\end{figure*}

As described in \cref{tab:apndx_cross_exp2}, fine-tuning across datasets from different devices (\eg, Samsung Galaxy Z-Fold 5 and ZTE Axon 20) improves generalization to different data distributions. Interestingly, the fine-tuned models $\text{DISCNet}_{s5z20}$ and $\text{UDC-UNet}_{s5z20}$ outperform DISCNet and UDC-UNet, solely trained by VidUDC33K, as shown in \cref{tab:apndx_cross_exp2}. This performance boost can be attributed to the fact that UDC-VIT exhibits more realistic and severe degradation patterns, such as noise, blur, transmittance decrease, and variant flares, compared to the synthetic VidUDC33K dataset, as discussed in \cref{sec:datasets_comparison} in the main body. Consequently, models pre-trained on UDC-VIT exhibit improved performance with fine-tuning on VidUDC33K, underscoring the benefits of utilizing real-world images as training datasets.

As described in the first row of \cref{fig:apndx_cross_exp2}(b), (c), and (d), models trained on UDC-VIT without fine-tuning (\eg, $\text{DISCNet}_{s5}$, $\text{UDC-UNet}_{s5}$, and $\text{DDRNet}_{s5}$) restore blur but fail to address flare artifacts. Interestingly, they show better restoration of transmittance decrease compared to VidUDC33K's ground truth, probably due to the brighter tone in UDC-VIT's ground truth compared to VidUDC33K's. On the other hand, models fine-tuned on VidUDC33K (\eg, $\text{DISCNet}_{s5z20}$, $\text{UDC-UNet}_{s5z20}$, and $\text{DDRNet}_{s5z20}$) effectively restore the complex degradation patterns specific to VidUDC33K, as shown in the second row of \cref{fig:apndx_cross_exp2}(b), (c), and (d).

\paragraph{Experiment 3: comparison of UDC-VIT and VidUDC33K.}

\begin{table*}[!b]
  \caption{\textbf{[Exp. 3]} Restoration performance of DDRNet~\cite{liu2024decoupling} trained on VidUDC33K~\cite{liu2024decoupling}, with or without additional fine-tuning on UDC-VIT. Models without subscripts refer to those trained solely on UDC-VIT, meaning their PSNR, SSIM, and LPIPS values match those in  \cref{tab:basic_experiment} in the main body. The number of iterations represents the percentage of fine-tuning iterations relative to the total iterations in the original configurations provided by the authors.}
  \label{tab:apndx_cross_exp3}
  \begin{center}
  \resizebox{0.8\linewidth}{!}{%
  \begin{tabular}{l|cccccc}
    \toprule
    Model name & PSNR~$\uparrow$ & SSIM~$\uparrow$ & LPIPS~$\downarrow$ & Training & Fine-tuning ($\# Iterations$) & Test \\
    \midrule
    $\text{DDRNet}_{z20}$   & 11.20 & 0.5331 & 0.5609 & VidUDC33K  &         -      & UDC-VIT \\
    $\text{DDRNet}_{z20s5}$ & 21.92 & 0.8302 & 0.2519 & VidUDC33K  & UDC-VIT ($10\%$) & UDC-VIT \\
    DDRNet         & 24.68 & 0.8539 & 0.2218 & UDC-VIT    &         -      & UDC-VIT \\
    \bottomrule
  \end{tabular}
  }
  \end{center}
\end{table*}

\begin{figure*}[!b]
\begin{minipage}{\linewidth}
   \centering
   \includegraphics[width=0.95\linewidth]{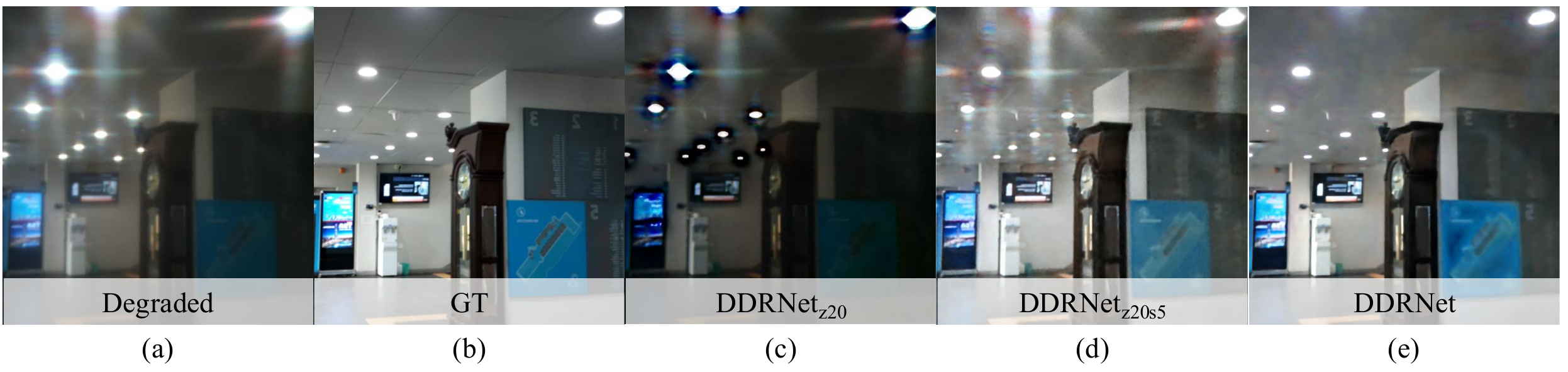}
   \caption{\textbf{[Exp. 3]} Restoration performance of DDRNet~\cite{liu2024decoupling} tested on the UDC-VIT dataset. (a) UDC-degraded and (b) GT images from the UDC-VIT dataset. Restored images by (c) $\text{DDRNet}_{z20}$, (d) $\text{DDRNet}_{z20s5}$, and (e) $\text{DDRNet}$. The model $\text{DDRNet}_{z20}$ is pre-trained on VidUDC33K without fine-tuning on UDC-VIT, while $\text{DDRNet}_{z20s5}$ is pre-trained on VidUDC33K with fine-tuning on UDC-VIT, showing improved restoration performance. However, compared to the results in \cref{fig:apndx_cross_exp2}, the fine-tuned model still struggles to handle the real-world degradations present in the UDC-VIT dataset, since it is pre-trained on the synthetic VidUDC33K dataset.}
\label{fig:apndx_cross_exp3}
\end{minipage}
\end{figure*}

This experiment evaluates the effect of fine-tuning on UDC-VIT when tested on UDC-VIT. We compare the performance of models trained on VidUDC33K, with or without fine-tuning on UDC-VIT, and models trained solely on UDC-VIT. For the models $\mathcal{M}$, we use DDRNet, the only publicly available pre-trained model trained on VidUDC33K. As shown in \cref{tab:apndx_cross_exp3}, $\text{DDRNet}_{z20}$ without fine-tuning on UDC-VIT fails to adequately handle the complex, severe, and real-world degradations present in UDC-VIT. In contrast, $\text{DDRNet}_{z20s5}$, fine-tuned on UDC-VIT, demonstrates significant performance improvements over $\text{DDRNet}_{z20}$. 

However, as illustrated in \cref{fig:apndx_cross_exp3}, even with fine-tuning, $\text{DDRNet}_{z20s5}$ still shows limitations in handling specific real-world degradations, such as severe flares. Notably, models pre-trained on the real-world UDC-VIT dataset and fine-tuned on the synthetic VidUDC33K dataset show strong performance in restoring degraded images in VidUDC33K (Experiment 2). In contrast, models pre-trained on the synthetic VidUDC33K dataset and fine-tuned on the real-world UDC-VIT dataset struggle to handle real-world degradation in UDC-VIT effectively (Experiment 3). These findings underscore the importance of pre-training on real-world datasets, such as UDC-VIT, to capture better complex degradations that synthetic data cannot fully replicate.

\section{Effects of Video Capturing Environment}
\label{sec:analysis}
\cref{tab:comparison_annotation} provides a comparative evaluation of the restoration performance across different deep-learning models, considering the impact of UDC-VIT dataset annotations. Since a single UDC-VIT video may include multiple annotations (\eg, an indoor scene with flares caused by artificial light), the annotation type listed in a column in \cref{tab:comparison_annotation} cannot be interpreted as the sole factor influencing UDC degradation. Nevertheless, it is reasonable to consider the annotation type as a key factor influencing PSNR, SSIM, and LPIPS values.

\begin{table*}[!b]
  \centering
  \caption{Comparison of restoration performance. Each row’s best and worst scores within each category are bold-faced and underlined, respectively.}
  \label{tab:comparison_annotation}
  \resizebox{1.0\linewidth}{!}{
  \begin{tabular}{l|c|ccc|c|c|cc|cc|c}
    \toprule
    \multirow{4}{*}{Model} & \multirow{4}{*}{Metric} & \multicolumn{5}{c|}{Flare presence and light sources} & \multicolumn{2}{c|}{Shooting location} & \multicolumn{2}{c|}{Human presence} & \multirow{4}{*}{Average} \\ \cline{3-11}
    &        & \multicolumn{4}{c|}{Present} & \multirow{3}{*}{Absent} & \multirow{3}{*}{Indoor} & \multirow{3}{*}{Outdoor} & \multirow{3}{*}{Present} & \multirow{3}{*}{Absent} & \\ \cline{3-6}
    &        & Natural & Artificial & \multirow{2}{*}{Both} & \multirow{2}{*}{Average} &  &  &  &  &  & \\
    &        & sunlight& light   &      &        &         &        &         &         &        &  \\
    \midrule
            & PSNR~$\uparrow$    & \underline{22.34}  & 24.15  & 22.37  & 23.64  & \textbf{26.92}  & \textbf{25.43}  & \underline{22.34}  & \textbf{26.94}  & \underline{20.33}  & 24.70 \\ 
    DISCNet~\cite{feng2021removing} & SSIM~$\uparrow$    & \underline{0.7704} & 0.8451 & 0.8191 & 0.8303 & \textbf{0.8611} & \textbf{0.8573 }& \underline{0.7852} & \textbf{0.8795} & \underline{0.7636}  & 0.8403 \\ 
            & LPIPS~$\downarrow$ & 0.2633 & 0.2894 & \underline{0.3250}  & 0.2901 & \textbf{0.2202} & \textbf{0.2608} & \underline{0.2891} & \textbf{0.2247} & \underline{0.3512} & 0.2675 \\ 
    \midrule
            & PSNR~$\uparrow$    & \underline{23.73}   & 27.76  & 25.36  & 26.83  & \textbf{30.46}  & \textbf{29.13}  & \underline{24.33}  & \textbf{31.34}  & \underline{21.47}  & 28.00 \\ 
    UDC-UNet~\cite{liu2022udc} & SSIM~$\uparrow$    & \underline{0.8197} & 0.8995 & 0.8857 & 0.8856 & \textbf{0.9027} & \textbf{0.9092} & \underline{0.8324} & \textbf{0.9276} & \underline{0.8198} & 0.8911 \\ 
           & LPIPS~$\downarrow$  & 0.1878 & 0.1814 & \underline{0.2173} & 0.1871  & \textbf{0.1588} & \textbf{0.1679} & \underline{0.2107} & \textbf{0.1398} & \underline{0.2526} & 0.1779 \\ 
    \midrule
               & PSNR~$\uparrow$ & 23.13   & 23.78  & \underline{21.49}  & 23.38  & \textbf{24.97}  & \textbf{24.34}  & \underline{22.45}   & \textbf{25.34}  & \underline{21.06}  & 23.89 \\ 
    FastDVDNet~\cite{tassano2020fastdvdnet} & SSIM~$\uparrow$ & \underline{0.7902} & 0.8523 & 0.8244 & 0.8392 & 0.\textbf{8538} & \textbf{0.8593} & \underline{0.7940} & \textbf{0.8720} & \underline{0.7891} & 0.8439 \\ 
            & LPIPS~$\downarrow$ & 0.2621 & 0.2772 & \underline{0.3048} & 0.2785 & \textbf{0.2403} & \textbf{0.2568} & \underline{0.2965} & \textbf{0.2364} & \underline{0.3244} & 0.2662 \\ 
    \midrule
            & PSNR~$\uparrow$    & 21.99  & 23.14  & \underline{21.57}  & 22.76  & \textbf{25.20}  & \textbf{24.07}  & \underline{21.88}  & \textbf{25.11}  & \underline{20.51}  & 23.55 \\ 
    EDVR~\cite{wang2019edvr} & SSIM~$\uparrow$       & \underline{0.7711} & 0.8422 & 0.8132 & 0.8276 & \textbf{0.8445}  &\textbf{ 0.8484} & \underline{0.7833}  & \textbf{0.8612} & \underline{0.7781} & 0.8331 \\ 
            & LPIPS~$\downarrow$ & 0.2565 & 0.2843 & \underline{0.3039} & 0.2826 & \textbf{0.2351} & \textbf{0.2605} & \underline{0.2893} & \textbf{0.2389}  & \underline{0.3227} & 0.2673 \\ 
    \midrule
            & PSNR~$\uparrow$    & \underline{23.61}  & 25.54  & 24.08  & 25.05  & \textbf{26.05}   & \textbf{26.07}  & \underline{23.12}   & \textbf{26.99}  & \underline{22.22}  & 25.38 \\ 
    ESTRNN~\cite{zhong2020efficient} & SSIM~$\uparrow$     & \underline{0.7997} & \textbf{0.8818} & 0.8577 & 0.8662 & 0.8637 & \textbf{0.8847} & \underline{0.8025} & \textbf{0.8938} & \underline{0.8098}  & 0.8654 \\ 
            & LPIPS~$\downarrow$ & 0.2415  & 0.2192 & \underline{0.2640}  & 0.2285 & \textbf{0.2072} & \textbf{0.2086} & \underline{0.2636} &\textbf{ 0.1920}  & \underline{0.2794} & 0.2216 \\ 
    \midrule
            & PSNR~$\uparrow$    & \underline{23.24}  & 24.14  & 23.49  & 23.92   & \textbf{26.25}  & \textbf{25.35}  & \underline{22.49}  & \textbf{26.43}  & \underline{21.23}  & 24.68 \\ 
    DDRNet~\cite{liu2024decoupling} & SSIM~$\uparrow$     & \underline{0.8015} & \textbf{0.8628} & 0.8455 & 0.8512 & 0.8594 & \textbf{0.8697} & \underline{0.8025}  & \textbf{0.8810}  & \underline{0.8007} & 0.8539 \\ 
            & LPIPS~$\downarrow$ & 0.2283 & 0.2267 & \underline{0.2433}& 0.2291 & \textbf{0.2066} & \textbf{0.2079} & \underline{0.2672} & \textbf{0.1936} & \underline{0.2771}  & 0.2218 \\ 
    \bottomrule
  \end{tabular}
  }
\end{table*}

\begin{figure*}[htbp]
   \centering
   \includegraphics[width=0.99\linewidth]{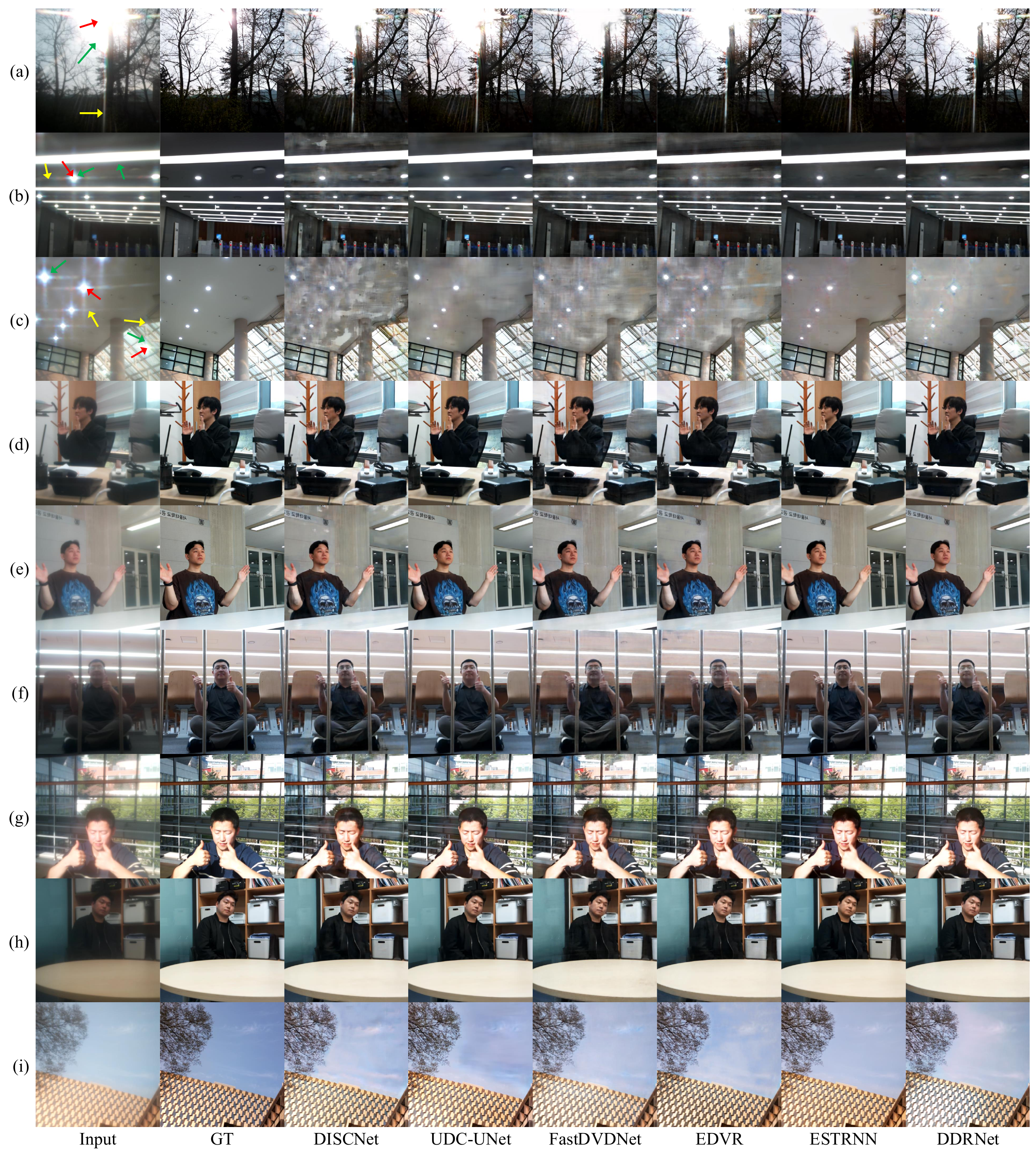}
   \caption{The visual comparison of the restoration performance regarding different annotations. The red, green, and yellow arrows represent the flares' glare, shimmer, and streak, respectively.
   (a) Natural sunlight + Human absent + Outdoor.
   (b) Artificial light + Human absent + Indoor.
   (c) Both + Human absent + Indoor.
   (d) Natural sunlight + Hand waving + Indoor.
   (e) Artificial light + Hand waving + Indoor.
   (f) Artificial light + Thumbs-up + Indoor.
   (g) Natural sunlight + Thumbs-up + Indoor.
   (h) Flare absent + Body-swaying + Indoor.
   (i) Flare absent + Human absent + Outdoor.
}
\label{fig:visual_comparison}
\end{figure*}

\paragraph{Light sources.}
As illustrated in \cref{fig:visual_comparison}, flares can be categorized into glare, shimmer, and streak~\cite{ahn2024udc,dai2022flare7k}, marked as red, green, and yellow arrows, respectively. A glare is characterized by intense and robust light, resulting in circular patterns as artifacts. Shimmer entails rapid and nuanced variations in light or color intensity across an image. A streak manifests as a lengthy, slender, and usually irregular line of light or color within an image.

As shown in \cref{tab:comparison_annotation}, all models encounter difficulties in restoring scenes with flare (Flare - Present - Average) compared to those without flare (Flare - Absent). Within flare-present scenes, the severity of degradation varies based on the light source (i.e., \textit{light source variant flare}) and location (i.e., \textit{spatially variant flare}). For example, in \cref{fig:visual_comparison}(a), sunlight-induced flares are intense, causing all models to struggle to restore obscured objects. In contrast, flares from artificial light, shown in \cref{fig:visual_comparison}(b) and (c), are relatively more straightforward to restore than sunlight-induced flares. Flares caused by natural light are often scattered by windows, as illustrated in \cref{fig:visual_comparison}(c) and (d), whereas those from artificial light sources with diffusers appear softened, as shown in \cref{fig:visual_comparison}(e). All models demonstrate effective restoration of these flare artifacts. However, some models fail to restore reflected light on the human face, as shown in \cref{fig:visual_comparison}(f), which may impact face recognition accuracy.

\paragraph{Shooting location.}
In the UDC-VIT dataset, 33.3\% of outdoor and 15.4\% of indoor scenes contain sunlight-induced flares. The benchmark models face more difficulty restoring outdoor scenes than indoor ones, as shown in \cref{tab:comparison_annotation}. Sunlight-induced flares in outdoor scenes (\cref{fig:visual_comparison}(a)) are more intense than those scattered by windows in indoor scenes (Figures~\ref{fig:visual_comparison}(c) and (d)).

Restoration performance is sometimes more affected by the presence of flares than by the shooting location alone. For example, among two indoor scenes, the models restore the flare-free frame (\cref{fig:visual_comparison}(h)) more effectively than the frame with flare (\cref{fig:visual_comparison}(b)). Similarly, among two outdoor scenes, the models perform better on the flare-free frame (\cref{fig:visual_comparison}(i)) than on the frame with flare (\cref{fig:visual_comparison}(a)).

\paragraph{Human.}
The presence of humans alone does not pose a significant challenge to restoration. Instead, the restoration difficulty hinges on how UDC degradations, such as noise, blur, transmittance decrease, and flare, impact humans. For example, in \cref{fig:visual_comparison}(d) and (e), despite the presence of flares in the frames, they do not affect humans. However, in \cref{fig:visual_comparison}(f), the reflection of fluorescent light on the person's glasses poses challenges to restoring fine details around the eyes. In \cref{fig:visual_comparison}(g) and (h), human faces appear reddish in the input frames compared to the ground-truth frames due to UDC-induced diffraction occurring differently across RGB channels. In addition, the restored facial colors vary among the models. Considering the importance of facial color in image and video-based applications, addressing diverse UDC degradations is essential for reliable human restoration in face recognition and video conferencing.

\paragraph{Flicker.}
The visual comparison presented in this paper is limited to a single frame. While specific frames in \cref{fig:visual_comparison} exhibit successful restoration, temporal inconsistencies, such as flickering, are frequently observed across multiple frames in the video for all models. This flickering may result from varying degradations between consecutive frames, such as decreases in transmittance and spatially and temporally variant flares. To view the restored videos featuring flickers, please visit \url{https://mcrl.github.io/UDC}.

\section{Details of the UDC-VIT Dataset}
\label{sec:add_info_udcvit}
In this section, we provide detailed information on the UDC-VIT dataset.

\subsection{Dataset Acquisition}
\label{sec:dataset_acquisition_apndx}
As described in \cref{sec:dataset_acquisition} in the main body of the paper, to construct a real-world UDC video dataset with precise alignment and synchronization, we propose a video-capturing system. We first describe the synchronization strategy between the two camera modules, which includes controller considerations, stable auto-exposure settings, fixed frame rates, and consistent pixel binning to ensure matched resolution. This section then details the alignment algorithm based on the Discrete Fourier Transform (DFT) and its advantages.

\paragraph{The controller.}
When capturing videos, we discard the initial 30 frames because it takes approximately 15 frames for the ground-truth camera and 25 frames for the UDC to achieve focus. The UDC requires more frames for focusing due to its degradation. Furthermore, we use a solid-state drive (SSD) instead of a secure digital (SD) card, since the SD card takes longer to save FHD resolution videos, which disrupts synchronization between the two cameras.

\paragraph{Exposure setting and synchronization.}
Since the presence of the display panel affects illuminance levels, all captures are performed in well-lit environments to ensure stable auto-exposure (AE). We fix the frame rate for temporal synchronization of corresponding frames in GT and degraded videos, while AE adjusts the exposure time. In such lighting conditions, both UDC and non-UDC cameras exhibit similar AE responses in terms of exposure time and gain. This is attributed to the QBC sensor’s high sensitivity, achieved through pixel binning that aggregates light from adjacent pixels. As a result, even with reduced transmittance from the display, the UDC camera achieves adequate brightness without significantly more prolonged exposure. In contrast, in low-light conditions (e.g., nighttime scenes), exposure mismatches can lead not only to synchronization issues but also to poorly rendered dark regions due to noise, underexposure, or loss of detail. Thus, we exclude such cases from the dataset.

\paragraph{Resolution consistency and image processing.}
To ensure consistent resolution between GT and degraded frames, we apply pixel binning consistently for both cameras using \texttt{rpicam} utility~\cite{rpicamApps}, constraining the output resolution below 16 megapixels (MP) (e.g., FHD). This binning process ensures that the same effective resolution is used across both clean and degraded captures. The output of this binning is then processed by the onboard ISP.

\paragraph{Alignment.}
The alignment algorithm we use involves \textit{shifting}, \textit{rotating}, and \textit{cropping} paired frames with DFT. The detailed algorithm is illustrated in \cref{alg:alignment}. In this algorithm, following Ahn~\etal~\cite{ahn2024udc}'s setting, we use $\lambda_1 = \lambda_3 = 1$ and $\lambda_2 = 0$, and we do not apply rotation. Their experiments show that applying rotation reduces the Percentage of Correct Keypoints (PCK) when varying $\lambda_1$, $\lambda_2$, $\lambda_3$, and $\theta_{\text{rotation}}$.

\begin{figure*}[!t]
   \centering
   \includegraphics[width=0.95\linewidth]{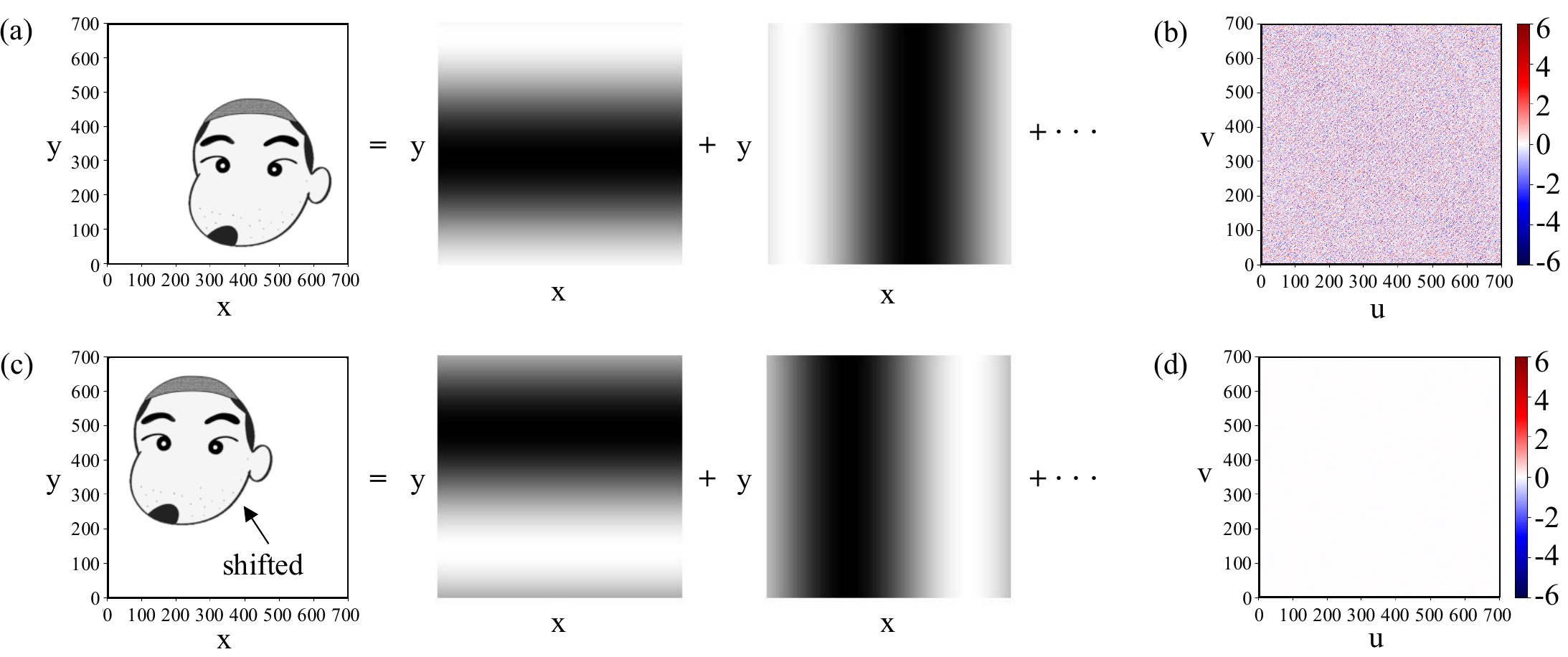}
   \caption{Frequency analysis based on the conceptual illustration for paired frames involving shifts without degradation. (a) The original frame $\mathcal{G}$ consists of multiple sinusoidal gratings. The inverse $DFT$ applied to $\mathcal{F}_{G}(u,v)$ produces each sinusoidal grating. (b) The phase difference between $\mathcal{G}$ and $\mathcal{D}$. (c) The spatially shifted frame $\mathcal{D}$ in the spatial domain comprises multiple sinusoidal gratings, as in (a). (d) The amplitude difference between $\mathcal{G}$ and $\mathcal{D}$, showing no difference.}
\label{fig:frequency_analysis}
\end{figure*}

The loss function in \cref{eq:loss} in the main body enables the incorporation of both local (i.e., MSE) and global (i.e., DFT) information across spatial and frequency domains. Using DFT to align the paired frames offers a significant advantage because it can decompose a frame into its constituent spatial frequency components. \cref{fig:frequency_analysis}(a) and (c) depict paired frames $\mathcal{G}$ and $\mathcal{D}$ comprising multiple sinusoidal gratings, indicating a noticeable spatial shift. \cref{fig:frequency_analysis}(b) and (d) represent the phase and amplitude differences between the paired frames, respectively. Thus, reducing the phase component is critical for effectively aligning the paired frames for the same scene.

\begin{algorithm*}[!b]
\caption{Alignment of paired images $I_G$ and $I_D$~\cite{ahn2024udc}.}
\label{alg:alignment}
\centering
\begin{minipage}{1.0\linewidth}
\begin{small}
\begin{algorithmic}
\Require Images $I_G$, $I_D$ of size $(H, W)$, hyperparameters $s$, $\theta_r$, $r$, $\lambda_1$, $\lambda_2$, $\lambda_3$
\Ensure Aligned images $\mathcal{G}$, $\mathcal{D}$ of size $(H^*, W^*)$
\State Crop $\mathcal{G}$ from $I_G$ using center crop
\State Crop $\mathcal{D}$ from $I_D$ to the size of $\mathcal{G}$
\State Initialize best loss $\mathcal{L}_{\text{best}}$ to a large value
\State Initialize optimal shifts $s_{\text{opt\_x}}$, $s_{\text{opt\_y}}$, and rotation $\theta_{\text{opt}}$ to 0

\For{$\theta_{\text{rotation}}$ from $-\theta_r$ to $\theta_r$ with step $r$}
    \State Apply rotation of $\theta_{\text{rotation}}$ to $I_D$ to get $\mathcal{D}_{\text{rotated}}$

    \For{$x_{\text{shift}}$ from $-s$ to $s$ with step 1}
        \For{$y_{\text{shift}}$ from $-s$ to $s$ with step 1}
            \State Calculate crop position $(p, q)$ relative to the center crop: \\
            \hspace{6em} $p = x_{\text{center\_crop}} + x_{\text{shift}}$ \\
            \hspace{6em} $q = y_{\text{center\_crop}} + y_{\text{shift}}$ 
            \State Crop image $\mathcal{D}_{\text{tmp}}$ from $\mathcal{D}_{\text{rotated}}$ at position $(p, q)$
            \State Calculate loss $\mathcal{L}$ using the loss function in \textbf{\cref{eq:loss}} between $\mathcal{D}_{\text{tmp}}$ and $\mathcal{G}$
            \If{$L < \mathcal{L}_{\text{best}}$}
                \State Update $\mathcal{L}_{\text{best}}$ to $L$
                \State Update $s_{\text{opt\_x}}$ to $x_{\text{shift}}$
                \State Update $s_{\text{opt\_y}}$ to $y_{\text{shift}}$
                \State Update $\theta_{\text{opt}}$ to $\theta_{\text{rotation}}$
            \EndIf
        \EndFor
    \EndFor
\EndFor

\State Apply optimal rotation $\theta_{\text{opt}}$ to $I_D$ to get $\mathcal{D}_{\text{rotated}}$
\State Calculate crop position $(p_{\text{opt}}, q_{\text{opt}})$ relative to the center crop: \\
\hspace{2em} $p_{\text{opt}} = x_{\text{center\_crop}} + s_{\text{opt\_x}}$ \\
\hspace{2em} $q_{\text{opt}} = y_{\text{center\_crop}} + s_{\text{opt\_y}}$
\State Crop $\mathcal{D}_{\text{rotated}}$ to acquire an aligned image $\mathcal{D}$ at position $(p_{\text{opt}}, q_{\text{opt}})$
\end{algorithmic}
\end{small}
\end{minipage}
\end{algorithm*}

\subsection{Dataset Details and Statistics}
This section provides detailed information about the UDC-VIT dataset.

\paragraph{Statistics.}
From a pool of 647 videos, we have randomly selected 510 for training, 69 for validation, and 68 for the test set. The UDC-VIT dataset is available in \texttt{PNG} format accompanied by a conversion script from \texttt{PNG} to \texttt{NPY}. We have annotated each video pair, providing a detailed overview of the total count and the distribution of different annotation labels. The video pairs are thoughtfully categorized into various conditions, including the presence of flare and light sources, the presence of humans and their motion types, and whether the scene is indoor or outdoor.

\begin{figure*}[!b]
   \centering
   \includegraphics[width=0.9\linewidth]{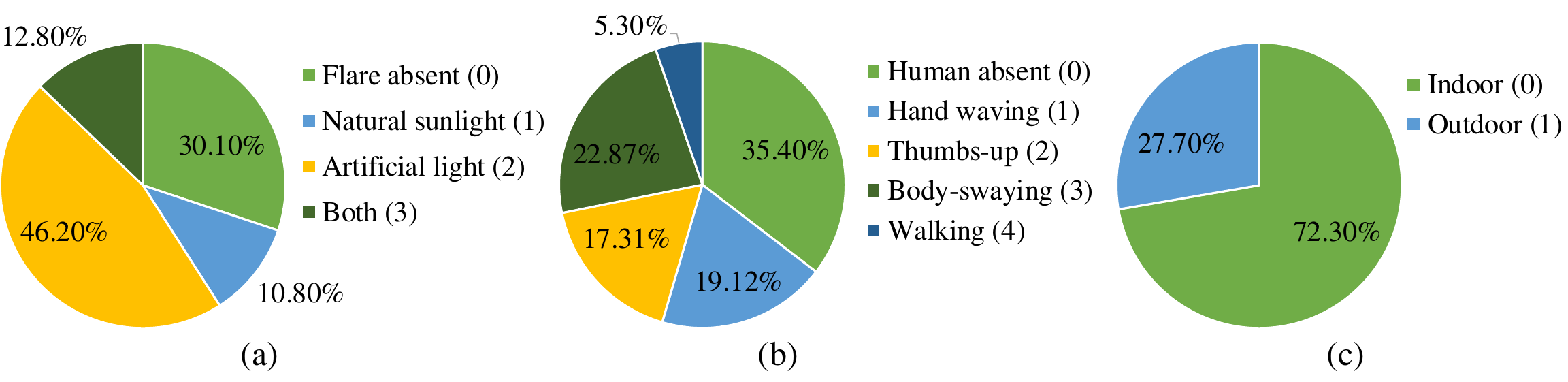}
   \caption{The dataset distribution. The parenthesis beside a label is the encoding of the label. Note that a video pair can have multiple annotation labels. The distribution of (a) lighting conditions, (b) human presence and their actions, and (c) shooting location.
   \label{fig:dataset_dist}
}
\end{figure*}

\begin{table*}[b]
  \centering
  \caption{Prescribed instructions from IRB-approved participant information sheet. Out of thirty shots per person, videos displaying issues such as being out-of-focus are eliminated from the dataset.}
  \begin{tabular}{p{0.97\linewidth}}
    \toprule
    \textbf{Q. What procedures will be followed if the participants take part in the study?} \\
    \midrule
    \textbf{A.} 
    Participants will undergo a 30-shot photo session using UDC and regular cameras with the following motions: \\
    \begin{itemize}
      \item 5-second shots of body-swaying $\times$ 9 shots (6 indoors / 3 outdoors)
      \item 5-second shots of waving hands $\times$ 9 shots (6 indoors / 3 outdoors)
      \item 5-second shots of giving a thumbs-up $\times$ 9 shots (6 indoors / 3 outdoors)
      \item 5-second shots of walking indoors/outdoors $\times$ 3 shots
    \end{itemize} \\
    Since the UDC is located beneath the display and operates in low-light environments, it is necessary to capture data in various locations (indoors and outdoors) and lighting conditions (bright and dark) to account for the diverse degradation patterns of the UDC. Additionally, recognizing individuals from different angles is crucial for tasks such as face recognition, especially in financial authentication. This requires deep-learning models capable of restoring a subject’s appearance from multiple perspectives (e.g., front, left, and right), and thus, a dataset with multi-angle captures is essential. The recorded videos will be publicly released as a dataset to support UDC research. \vspace{0.5\baselineskip} \\ 
    \toprule
    \textbf{Q. How long will the study participation last?} \\
    \midrule
    \textbf{A.} The study will take approximately 30 minutes. While the actual recording will take 2 minutes and 30 seconds (5 seconds $\times$ 30 shots), additional time will be needed for: \\
    \begin{itemize}
      \item Adjusting the subject's shooting angle (5 minutes)
      \item Transitioning between locations (5 minutes)
      \item Verifying alignment accuracy (5 minutes)
      \item Conducting necessary adjustments (10 minutes)
    \end{itemize} \\
    \toprule
    \textbf{Q. Will compensation be provided for participating in this study?} \\
    \midrule
    \textbf{A.} As a token of appreciation, participants who complete all 30 video captures will receive a \$40 Starbucks gift card. However, participants who withdraw before completing all 30 shots or request video deletion will not be compensated. If a participant requests video deletion after receiving compensation, they must return the full amount. \vspace{0.5\baselineskip} \\
    \bottomrule
  \end{tabular}
  \label{tab:irb_info_sheet}
\end{table*}

\paragraph{IRB approval.}
We have obtained the approval of the Institutional Review Board (IRB) for the UDC-VIT dataset, as our research involves human subjects. This rigorous process ensures the highest standards of research ethics. Using IRB-approved procedures, we enlisted 22 voluntary research participants. As shown in \cref{tab:irb_info_sheet}, the IRB-approved information sheet provides detailed instructions to the participants.

Similarly, it is essential to note that the users of the UDC-VIT dataset are engaged in research involving human subjects. Therefore, users are required to obtain IRB approval by the regulations of their respective countries. When users download the dataset, instructions on IRB approval will be provided, as shown in \cref{fig:thunder_homepage}.

\subsection{Rigorous Maintenance Plan}
This section provides easy accessibility and a rigorous maintenance plan for the UDC-VIT dataset.

\begin{figure*}[!b]
   \centering
   \includegraphics[width=0.95\linewidth]{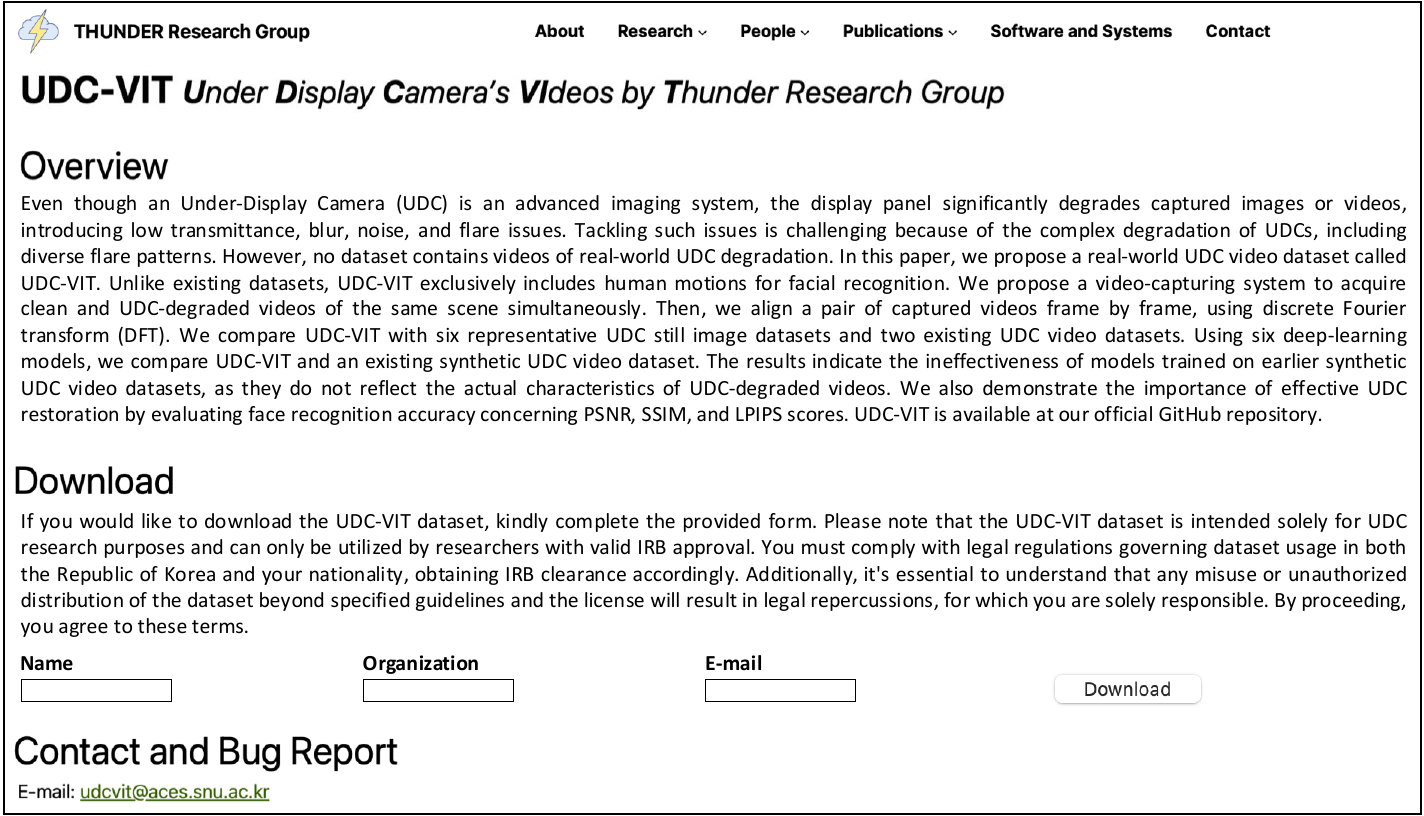}
   \caption{Our research group's official website providing detailed information and a download link for the UDC-VIT dataset.}
\label{fig:thunder_homepage}
\end{figure*}

\paragraph{Easy accessibility.}
The UDC-VIT dataset will be publicly available on our research group's website, as shown in \cref{fig:thunder_homepage}. Users can request access by completing a form on the site; a download link will be sent via email upon submission. Detailed instructions are available at \url{https://github.com/mcrl/UDC-VIT}. Hosting the dataset on the research group's website ensures long-term availability, while handling contact and bug reports via email facilitates ongoing maintenance and updates.

\paragraph{License.}
The UDC-VIT dataset is licensed under Creative Commons Attribution-NonCommercial-ShareAlike 4.0 International (CC BY-NC-SA 4.0). Under this license, users of the UDC-VIT dataset can freely utilize, share, and modify this work by adequately attributing the original author, distributing any derived works under the same license, and utilizing it exclusively for non-commercial purposes. It is essential to mention that the UDC-VIT dataset is restricted to UDC research purposes only, as outlined in our IRB documentation. Detailed information on this license can be found on the official Creative Commons website.


\section{Reproducibility}
\label{sec:apndx_reproducibility}
This section provides detailed information on the deep-learning models for reproducibility. The code can be found and downloaded at \url{https://github.com/mcrl/UDC-VIT}.

The learnable restoration models used for evaluating the UDC-VIT dataset include DISCNet~\cite{feng2021removing}, UDC-UNet~\cite{liu2022udc}, FastDVDNet~\cite{tassano2020fastdvdnet}, EDVR~\cite{wang2019edvr}, ESTRNN~\cite{zhong2020efficient}, and DDRNet~\cite{liu2024decoupling}. We use a single-node GPU cluster to train each benchmark model. Each node has eight AMD Instinct MI100 GPUs. While we mainly stick to the original authors' code and training settings for the models, we introduce some modifications to all models except ESTRNN.

\begin{itemize}

\item \textbf{DISCNet.}
DISCNet is designed to restore UDC still images in high dynamic range (HDR) (\eg, SYNTH~\cite{feng2021removing}). Accordingly, we modify the PyTorch DataLoader to use normalization instead of Reinhard tone mapping~\cite{reinhard2002photographic}. The DataLoader randomly selects one frame per video from the UDC-VIT dataset for each iteration during the training and validation phases.

\item \textbf{UDC-UNet.}
UDC-UNet is designed to restore UDC still images in HDR. The original authors do not conduct normalization or tone mapping in the DataLoader and employ a tone mapping L1 loss function. However, since the UDC-VIT dataset has a low dynamic range (LDR), we modify the DataLoader to normalize the input. The model output is clamped between 0 and 1 before computing the L1 loss. The DataLoader randomly selects one frame per video from the UDC-VIT dataset for each iteration.

\item \textbf{FastDVDNet.}
FastDVDNet is a video denoising model that utilizes NVIDIA's Data Loading Library (DALI)~\cite{nvidia2018dali}, processing a noise map and multiple frames as inputs. Instead of DALI, we employ the PyTorch DataLoader tailored to the UDC-VIT dataset in \texttt{NPY} format. We set the noise level to zero. To accommodate the FHD resolution and multiple degradations in the UDC-VIT dataset, we increase the patch size from 64 to 256. Moreover, we extend the training duration of FastDVDNet from the original 95 epochs to 400, allowing the model to reach full convergence.

\item \textbf{EDVR.}
To address the out-of-memory issues of EDVR, which contains 23.6M parameters, we reduce the training patch size from 256 to 192. During inference on the test set, we divide each frame into two patches of size $3 \times 1060 \times 1060$ and merge the outputs afterward. 

\item \textbf{DDRNet.}
During inference, the authors of DDRNet partition each frame into patches of size $3 \times 256 \times 256$ and process 50 frames simultaneously. However, this patch-wise inference introduces visible seams at the patch boundaries. To mitigate this artifact, we instead perform inference at full resolution ($3 \times 1060 \times 1900$), using ten consecutive frames at a time.

\end{itemize}

\section{Discussion on the Responsible Use of the Dataset}
\label{sec:discussion}
This section discusses potential negative societal impacts, corresponding user guidelines, and our responsibilities.

\subsection{Potential Negative Societal Impacts}
The UDC-VIT dataset contains facial images and motions of 22 research participants, which raises concerns about potential misuse, such as in deepfake applications. This technology can generate convincingly altered videos, posing risks to individual privacy and societal trust. Deepfakes may infringe upon personal integrity and privacy, potentially leading to social unrest and confusion. Given these potential negative societal impacts, careful consideration is required when using the dataset.

\subsection{User Guidelines}
Users of the UDC-VIT dataset are expected to adhere to the following guidelines:

\begin{itemize}

\item \textbf{Responsible use.}
Users must utilize the dataset ethically and responsibly, ensuring that its use does not cause societal harm or infringe on individual privacy.

\item \textbf{Compliance with legal and ethical standards.}
Users must comply with all applicable legal and ethical standards. This includes obtaining approval from an Institutional Review Board (IRB) by the regulations of the respective country and adhering to any restrictions or conditions imposed by the IRB or other regulatory bodies. Any legal violations will be the sole responsibility of the user.

\item \textbf{Restricted usage.}
The UDC-VIT dataset must not be used for harmful applications, such as deep fake technologies or the creation of misinformation and manipulation. Furthermore, the 22 participants agreed to a research scope defined during our IRB review, which is limited to acquiring UDC video data and developing restoration models. Therefore, the dataset must be used exclusively for UDC-related research purposes.

\end{itemize}

\subsection{Our Responsibility}
As the administrators of the UDC-VIT dataset, we acknowledge our responsibility to:

\begin{itemize}

\item \textbf{Protect participant privacy.}
Our primary priority is to maintain the privacy and confidentiality of our research participants. Although the participants provided consent for the public use of their facial appearances and movements within the dataset, we are committed to guiding users toward appropriate research practices and to making efforts to safeguard any additional personal information.

\item \textbf{Facilitate ethical use.}
We provide comprehensive documentation and ethical usage guidelines through the \textit{Datasheets for Datasets} and our GitHub repository at \url{https://github.com/mcrl/UDC-VIT}. The dataset can be accessed via an automatic email system after users complete the application form on our research group's website. This process also informs users of potential risks and ethical considerations associated with the dataset.

\item \textbf{Respond to concerns.}
We are committed to the responsible stewardship of the UDC-VIT dataset and will respond promptly to any concerns or complaints that may arise. We value user feedback and are prepared to take appropriate actions—such as data correction or updates—to prevent or mitigate harm, should any misuse of the dataset be reported through our research group’s website (see \cref{fig:thunder_homepage}).

\end{itemize}

\end{document}